%% file: root.tex
\documentclass[letterpaper, 10 pt, journal, twoside]{IEEEtran}

\IEEEoverridecommandlockouts  

\usepackage{hyperref}

\usepackage{multibib}
\newcites{suppl}{Supplementary References}

\usepackage{datetime2}
\DTMsetup{
	datesep={},
	timesep={}
}

\usepackage{graphics} 
\usepackage{amsmath} 
\usepackage{amssymb}  
\usepackage{booktabs}
\usepackage[font={footnotesize}, skip=5pt]{caption}
\usepackage{subcaption}
\usepackage{microtype}
\usepackage{flushend}
\usepackage{cuted}
\let\labelindent\relax
\usepackage[inline]{enumitem}
\renewcommand{\baselinestretch}{0.99}
\usepackage{diagbox}

\usepackage[T1]{fontenc}

\usepackage{graphicx}

\graphicspath{{figures/}}

\newcommand{\secref}[1]{Sec.~\ref{#1}}
\renewcommand{\eqref}[1]{Eq.~(\ref{#1})}
\newcommand{\figref}[1]{Fig.~\ref{#1}}
\newcommand{\tabref}[1]{Tab.~\ref{#1}}

\newcommand\rebuttal[1]{#1}  

\usepackage[noadjust]{cite}

\usepackage{pgfplots}
\pgfplotsset{compat=1.18}

\usepackage[colorinlistoftodos,prependcaption]{todonotes}

\usepackage{siunitx}

\newcommand\independent{\protect\mathpalette{\protect\independenT}{\perp}}
\def\independenT#1#2{\mathrel{\rlap{$#1#2$}\mkern2mu{#1#2}}} 

\DeclareMathOperator*{\argmax}{\arg\!\max}

\usepackage[abs,unit=1mm]{overpic}

\usepackage{csquotes}

\usepackage{float}
\newlength{\imagewidth}

\newcommand\subfiguresubref[1]{Subfig.~\subref{#1}}

\newsavebox{\tempbox}

\usepackage{xspace} 
\newcommand{\closemicrowave}{\texttt{CloseMicrowave}\xspace}
\newcommand{\takelidoffsaucepan}{\texttt{TakeLidOffSaucepan}\xspace}
\newcommand{\phoneonbase}{\texttt{PhoneOnBase}\xspace}
\newcommand{\putrubbishinbin}{\texttt{PutRubbishInBin}\xspace}
\newcommand{\pickup}{\texttt{PickUp}\xspace}
\newcommand{\pushtogoal}{\texttt{PushToGoal}\xspace}
\newcommand{\pickandplace}{\texttt{PickAndPlace}\xspace}

\title{\LARGE \bf
The Treachery of Images: Bayesian Scene Keypoints\\for Deep Policy Learning in Robotic Manipulation}

 \author{Jan Ole von Hartz$^{1}$, Eugenio Chisari$^{1}$, Tim Welschehold$^{1}$, Wolfram Burgard$^{2}$,\\ Joschka Boedecker$^{1}$ and Abhinav Valada$^{1}$
 \thanks{$^{1}$Department of Computer Science, University of Freiburg, Germany.}
 \thanks{$^{2}$Department of Engineering, University of Technology Nuremberg.}
 \thanks{This work was funded by the BrainLinks-BrainTools center of the University of Freiburg, the Carl Zeiss Foundation with the ReScaLe project, and by an academic grant from NVIDIA. (\emph{Corresponding author: Jan Ole von Hartz} \href{mailto:hartzj@cs.uni-freiburg.de}{hartzj@cs.uni-freiburg.de})}
 }

\let\oldtwocolumn\twocolumn
\renewcommand\twocolumn[1][]{%
    \oldtwocolumn[{#1}{
    \begin{center}
        \includegraphics[width=\textwidth]{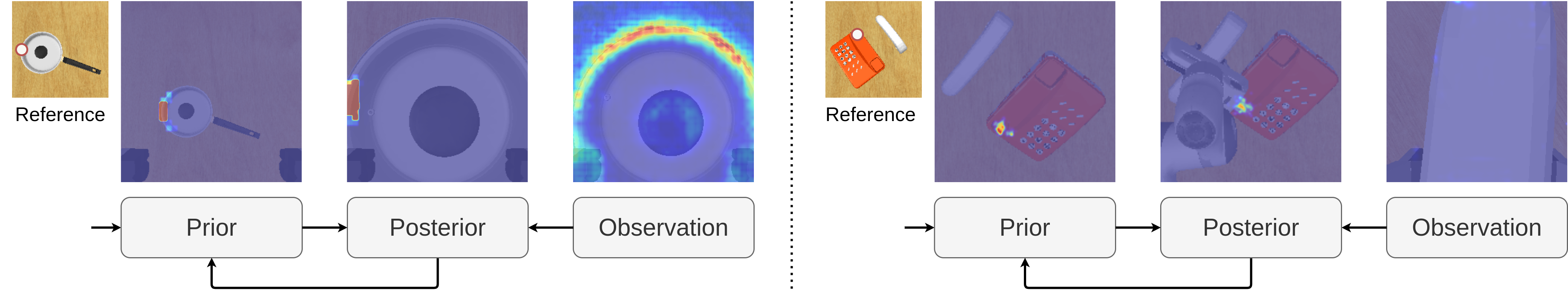}
        \captionof{figure}{%
       Individual camera observations are often ambiguous.
       For example, from the observation on the left, the rotation of the saucepan cannot be uniquely inferred.
       When tracking object keypoints, this leads to \rebuttal{statistically} multimodal localization hypotheses.
       We overcome this problem by considering the image in context.
       We find likely correspondences across image scales and then use spatial or temporal context to resolve the ambiguities.
       Our model further detects when a keypoint is likely not observed, enabling our approach to track occluded objects and objects outside the current field of view as shown on the right.}
       \label{fig:filter_flow}
    \end{center}
    }]
}

\begin{document}
  
\maketitle
\thispagestyle{empty}
\pagestyle{empty}

\begin{abstract}
        In policy learning for robotic manipulation, sample efficiency is of paramount importance.
        Thus, learning and extracting more compact representations from camera observations is a promising avenue.
        However, current methods often assume full observability of the scene and struggle with scale invariance.
        In many tasks and settings, this assumption does not hold as objects in the scene are often occluded or lie outside the field of view of the camera, rendering the camera observation ambiguous with regard to their location.
        To tackle this problem, we present BASK, a Bayesian approach to tracking scale-invariant keypoints over time.
        Our approach successfully resolves inherent ambiguities in images, enabling keypoint tracking on symmetrical objects and occluded and out-of-view objects.
        We employ our method to learn challenging multi-object robot manipulation tasks from wrist camera observations and demonstrate superior utility for policy learning compared to other representation learning techniques.
        Furthermore, we show outstanding robustness towards disturbances such as clutter, occlusions, and noisy depth measurements, as well as generalization to unseen objects both in simulation and real-world robotic experiments.
\end{abstract}

\input{sections/introduction.tex}
\input{sections/related_work.tex}
\input{sections/approach.tex}
\input{sections/experiments.tex}
\input{sections/conclusion.tex}


\typeout{}
\footnotesize
\bibliographystyle{IEEEtran}
\bibliography{root}

\input{sections/supplementary}

\end{document}

%% file: sections/introduction.tex
\section{Introduction}

Over the last decade, policy learning methods that learn their own visual representations end-to-end have become exceedingly popular~\cite{mnih2013playing, chisari2022correct, burgard2020perspectives}.
These methods are helpful in robotics, where we rarely have access to ground truth scene features but have to learn from raw camera observations instead.
However, learning environment features from scratch can be prohibitively expensive, with current approaches often requiring large amounts of training data. Using pretrained visual models improves the efficacy of policy learning in robot manipulation~\cite{wulfmeier2021representation, schmalstieg2023learning, younes2023catch}.
Nevertheless, current approaches still fail on complex tasks, and no representation has been able to fully close the gap  to models learning on ground truth scene features.

This is primarily due to three fundamental problems.
First, efficient expert cognition is goal-directed~\cite{drew2013invisible}.
Thus, a pretraining objective that requires semantic image understanding is required for better downstream policy success~\cite{florence2019self}. 
Second, state-of-the-art representation learning methods find it challenging to find corresponding visual features across different image scales~\cite{graf2022learning}.
Consequently, it is difficult to employ them with wrist-mounted cameras, whereas such cameras are widely available in real-world settings and enable many robotic tasks~\cite{hsu_vision-based_2022, honerkamp2022n}.
Finally, images are treated in isolation and \rebuttal{scene} context is neglected.
Although images are highly ambiguous, for instance, due to object symmetries and occlusions.
These ambiguities can only be resolved in \rebuttal{temporal or spatial} context, as shown in \figref{fig:filter_flow}.
Similarly, any representation that we derive from an ambiguous image is prone also to be ambiguous.
Hence, to generate an unambiguous image representation, the integration of context is essential.
Moreover, due to their neglect of context, current methods cannot represent occluded objects or objects outside the field of view which further limits their applicability.

To address the these problems, we present \textbf{Ba}yesian \textbf{S}cene \textbf{K}eypoints (BASK), a novel approach that focuses on representing the underlying scene, instead of representing each image in isolation.
We address the first problem by extracting goal-directed information from each image via localizing 3D scene keypoints using a semantic encoder network and adding depth information.
To address the second problem, we pay special attention to corresponding visual features across scales while training the network.
Finally, we integrate our localization hypotheses across time and camera views to resolve multi-modalities and represent temporarily unobserved objects.
We propose an approach to integrating observations based on the Bayes filter, which makes minimal and transparent assumptions.

We extensively evaluate the efficacy of our approach for learning challenging multi-object manipulation policies. We compare against a suite of other representation methods using RLBench~\cite{james2019rlbench}, a standard benchmark of manipulation tasks involving everyday objects. We establish superior efficacy for policy learning and improved localization accuracy, especially when learning from wrist camera observations.
Further confirming the efficacy of our method in real-world experiments, we find our approach to transfer significantly better to the complexities of real-world perception than other methods.
We further observe zero-shot transfer to cluttered scenes and previously unseen objects and environments.
Enabling efficient policy learning from wrist camera observations, our method frees policy learning approaches from the confines of a lab where object tracking systems and camera arrays are available.
Thus, it enables a plethora of applications such as mobile manipulation and the deployment of robots in environments where overhead cameras viewing the entire workspace are not available.

In summary, our main contributions are:
\begin{enumerate}
    \item We propose a new framework for representation learning, interpreting it as \emph{scene} representation instead of image representation. In this framework, we develop a Bayesian approach to resolving the inherent ambiguities of images.
    \item We train encoder networks to generate semantic descriptions of multi-object scenes, respecting scale variances and occlusions, e.g.\ due to moving cameras, and leverage our Bayes filter to resolve the resulting ambiguities.
    \item We rigorously evaluate the ability of our approach to overcome the problems outlined above, as well as its efficacy for policy learning, both in simulation and real-world robotic experiments.
    \item We make the code and models publicly available at \url{http://bask.cs.uni-freiburg.de}.
\end{enumerate}

The supplementary material is appended at the end of this paper.

%% file: sections/related_work.tex
\section{Related Work}\label{sec:related_work}
Our goal is to generate compact representations from camera observations suited for efficient policy learning. 
To be applicable for wrist camera observations, these should be scale and occlusion invariant and be able to represent objects that are temporarily occluded or outside the field of view.
Existing representation learning methods range from implicit methods, neurally compressing the image~\cite{higgins2017beta, burgess2019monet} to methods explicitly expressing the poses of scene objects~\cite{deng2009imagenet, wang2019densefusion}.
Our keypoints-based approach lies in the middle of this spectrum, thus avoiding the major drawbacks of either family of methods.

{\parskip=2pt
\noindent\textit{Implicit methods}: 
A common approach to compress camera observations is training a neural network with a bottleneck on image reconstruction, such as using Variational Auto-Encoders (VAEs).
\rebuttal{Instances of this approach are \(\beta\)-VAE~\cite{higgins2017beta}, which can produce disentangled representations, and MONet~\cite{burgess2019monet}, which partitions the image into several \emph{slots} first, thus separating objects.}
Similarly, Transporter~\cite{kulkarni2019unsupervised} is trained to reconstruct a source image from a target image via transporting local features.
These representations have been shown to enable more efficient policy learning on a set of robotic manipulation tasks~\cite{wulfmeier2021representation}.
\rebuttal{Making few assumptions, they are flexibly applicable.}
However, they are not as effective for downstream policy learning as ground truth scene features~\cite{wulfmeier2021representation} or object keypoints~\cite{florence2019self}.
}

{\parskip=2pt
\noindent\textit{Explicit methods}: 
Pose estimation methods~\cite{deng2019poserbpf, wang2019densefusion, piga2021maskukf} represent the pose of the relevant scene objects explicitly.
However, they typically need a ground truth 3D model of the object~\cite{deng2019poserbpf, piga2021maskukf}, they are not applicable to deformable objects~\cite{deng2019poserbpf, piga2021maskukf} and do not work well in the presence of occlusions.
Recent methods that do not require any CAD models or object-specific training~\cite{sun2022onepose, liu2022gen6d}, still require a pre-recorded scan of the object of interest for inference and are only applicable to rigid objects.
More importantly, in \secref{sec:pose_est} in the supplementary material we show that they do not exhibit the needed scale-invariance for learning from a wrist camera.
}

{\parskip=2pt
\noindent\textit{Keypoints}:
Keypoints are pixel- or 3D coordinates tracking task-relevant object parts.
KETO~\cite{qin2020keto} predicts single keypoints from 3D object point clouds, whereas Neural Descriptor Fields~\cite{simeonov2022neural} encode the full object point cloud.  
However, extracting full object point clouds is challenging and requires an array of cameras surrounding the scene, limiting the applicability of the method.
\rebuttal{Keypoints can be learned end-to-end from camera observations in RL~\cite{chen2021unsupervised} or using multi-view consistency~\cite{vecerik2021s3k}.}
In their seminal work, Florence~\textit{et~al.}~\cite{florencemanuelli2018dense, florence2019self} generate keypoints by training Dense Object Nets (DON)
in a self-supervised manner.
DONs can generalize between object class instances~\cite{florencemanuelli2018dense} and are applicable to deformable objects~\cite{florence2019self}.
Training DONs to contrast between different single-object scenes, enables deployment in a multi-object setting~\cite{hadjivelichkov2022fully}.
However, this approach requires single-object scans of all objects and handles occlusions poorly.
In contrast, we propose to directly train on multi-object scenes, making data collection much faster, avoiding computational overhead and including occlusions in the training data.
Moreover, in densely cluttered scenes, the object mask generation needed for DON training can be skipped~\cite{adrian2022efficient, graf2022learning} 
However, on less cluttered scenes this approach runs the risk of sampling too many background pixels in pretraining, thus compromising correspondence quality.

\begin{figure*}[tb]
    \includegraphics[width=\textwidth]{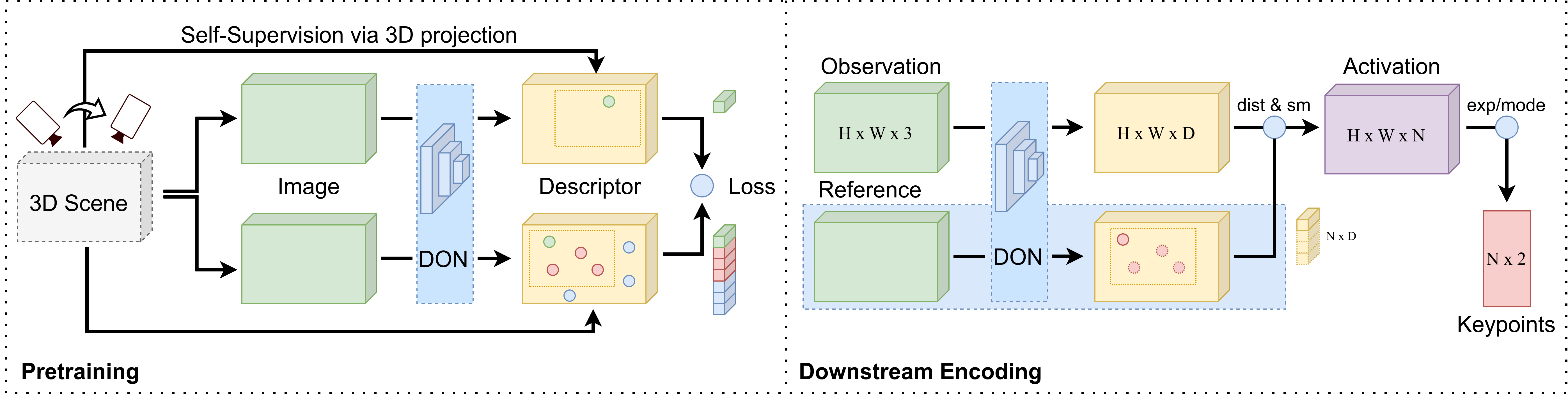}
    \caption{Pretraining and keypoint generation processes of dense object nets. The encoder is pretrained using a self-supervised pixel-wise contrastive loss which optimizes the descriptor distance on scans of static scenes. During downstream policy learning, the descriptor of the current observation is compared to a previously selected set of reference descriptors and the pixel coordinates of the respective most likely match are used as the location of the keypoint.}\label{fig:don_flow}
    \vspace{-0.3cm}
\end{figure*}

A major open problem in keypoint detection is visual correspondence across image scales.
DONs fail to find accurate correspondences for vertical camera translations of less than \SI{10}{\cm}~\cite{graf2022learning}, rendering wrist cameras less effective for policy learning.
Prior work circumvents this problem by corresponding from a fixed height~\cite{florence2019self, graf2022learning, florencemanuelli2018dense,hadjivelichkov2022fully, adrian2022efficient}.
Similarly, prior work does not comprehensively address the problem of occlusion~\cite{florencemanuelli2018dense, florence2019self}.
We demonstrate that DONs can be trained to be invariant to the image scale and occlusions, while distinguishing between multiple objects.
Specifically, we find multi-model hypotheses to emerge which we resolve using Bayesian inference.
}

{\parskip=2pt
\noindent\textit{\rebuttal{Visual Tracking}}: 
Ample methods have been developed to disambiguate local visual features within the context of a single image~\cite{han2005bayesian, oliver2000bayesian, sarlin2020superglue}.
In contrast, we consider images that are inherently ambiguous due to occlusions or otherwise missing \rebuttal{scene} context and leverage context to resolve the ambiguities.
\rebuttal{Correlation filters can be used to track visual features across time~\cite{ma2015long}, but they cannot utilize 3D information and the semantic generalization of DONs.
Moreover, their tracker needs to be specifically trained, whereas our Bayes filter works off-the-shelf.
Similarly, Bayesian approaches for tracking visual features across videos~\cite{zhao2004tracking, kwon2010visual} do not use 3D information and do not generalize across object instances.}
To the best of our knowledge, we are the first to leverage the potential of Bayes Filter in representation learning for robotic manipulation.}

%% file: sections/approach.tex
\section{Technical Approach}

We aim to generate 3D keypoints as an efficient representation for downstream policy learning.
To be applicable to wrist camera observations, these keypoints need to be scale and occlusion invariant.
Furthermore, they should be able to track multiple relevant scene objects and represent objects that are temporally occluded or outside the camera's field of view.
Our approach, \textbf{Ba}yesian \textbf{S}cene \textbf{K}eypoints (BASK), is two-pronged.
First, we find semantic correspondences between images.
To this end, we train Dense Object Nets (DON) directly on multi-object scenes. 
\rebuttal{We compute the localization hypotheses for a keypoint by comparing the corresponding reference descriptor to the descriptor image generated by the DON.}
Ambiguous images lead to multimodal hypotheses.
We then integrate these hypotheses using the Bayes filter to resolve ambiguities.

\input{sections/keypoints.tex}
\input{sections/scene_representation.tex}

%% file: sections/keypoints.tex
\subsection{Learning Semantic Correspondence}\label{sec:kp}
To extract keypoints from camera observations, we train a DON in a self-supervised manner~\cite{florencemanuelli2018dense} and use the generated embeddings for downstream keypoint generation~\cite{florence2019self}.
We then adapt these techniques to the multi-object case and describe how to achieve invariance towards scale, rotation, and occlusions.

\subsubsection{Dense-Correspondence Pretraining}\label{sec:pretrain}
By moving an RGB-D camera in a static scene and tracking the camera pose, we reconstruct the 3D representation of that scene using volumetric reconstruction.
After filtering out background points, we project the object point cloud back onto the image plane to generate object masks for all images along the trajectory.
For a given pixel position in one image in the trajectory, we find the corresponding pixel position in another image of the same trajectory via simple 3D projections, using the respective camera pose and calibration matrix.

Using this technique for finding correspondences between pairs of images, we train an encoder network \(e_\eta:\mathbb{R}^{H\times W\times 3}\to\mathbb{R}^{H\times W\times D}\), mapping an RGB image to a \(D\)-dimensional descriptor, to minimize the descriptor distance between corresponding points while enforcing at least a margin \(M\) between non-corresponding points.
Specifically, for a given pair of images \(I_a, I_b\), we sample a set of \rebuttal{\(m\)} pixel locations \(U_a\) from the object mask of \(I_a\) and compute the set of corresponding pixel positions \(U_b\) \(I_b\).
Additionally, for each point \(u_a\in U_a\) we sample a set of \rebuttal{\(n\)} non-corresponding points \(N_{u_a}\) from both \(I_b\)'s object mask and the background.
\rebuttal{Let \(e_\eta(I_a)_{u_a}\) denote the value of descriptor image \(e_\eta(I_a)\) at position \(u_a\).}
We then compute \rebuttal{the loss} for the encoder \rebuttal{\(e_\eta\)} as
\begin{align}
    \mathcal{L}&(I_a, I_b) = \sum_{u_a, u_b\in U_a, U_b} \left( \frac{\Vert e_\eta(I_a)_{u_a} - e_\eta(I_b)_{u_b}\Vert^2}{m} \right.\nonumber\\
    &\ \left. + \sum_{u_c\in N_{u_a}}  \frac{\max\left(0, M-\Vert e_\eta(I_a)_{u_a} - e_\eta(I_b)_{u_c}\Vert^2\right)}{n} \right).
\end{align}\label{eq:dc_loss}
 
We found improved correspondence quality by enforcing a larger margin \(M_{\mathit{bg}}\) for background non-matches than for the foreground non-matches \(M_{\mathit{fg}}\).
\figref{fig:don_flow} illustrates this approach.

\subsubsection{Multi-Object Tasks}\label{sec:mo}
To extend DONs to multi-object tasks, we directly train on multi-object scenes such that the data is fast to collect and already contains occlusions. 
We again employ volumetric reconstruction and split the resulting point cloud using density-based clustering~\cite{ester1996density}.
Projecting these object-wise point clouds back onto the camera planes yields consistent object masks for the trajectory.
Furthermore, when working with a multi-camera setup, we can generate masks that respect occlusions, e.g.\ caused by the robot arm, further diversifying the training data.
During one step of pretraining, we sample one of the object masks and treat the other objects as part of the background.
This ensures that the model learns to distinguish the different objects.

\subsubsection{Invariances}\label{sec:inv}
DONs struggle to generalize to camera perspectives outside the training distribution, especially for vertical camera movements~\cite{graf2022learning}.
This is not limited to cases where the change in perspective removes necessary \rebuttal{scene} context but to changes in distance between object and camera in general.

\paragraph{\rebuttal{Scale}}
We hypothesize that this is due to CNNs being biased towards texture~\cite{geirhos2018imagenet, naseer2021intriguing}, with textures transferring badly between image scales due to image rasterization occurring at a fixed resolution.
Hence, correspondences across scales can only emerge at a higher semantic feature level.
However, scale invariance is a critical property to be able to use DONs on wrist camera observations.
By training the network on multiple image scales, we teach it to generate a more scale-invariant and semantically meaningful representation.
Moreover, in contrast to previous work~\cite{florence2019self}, we automatically scale the fraction of masked non-matches by the relative size of the object mask in the image to better account for scale differences.

\begin{figure}[tb]
        \centering
        \settowidth{\imagewidth}{\includegraphics[scale=.07]{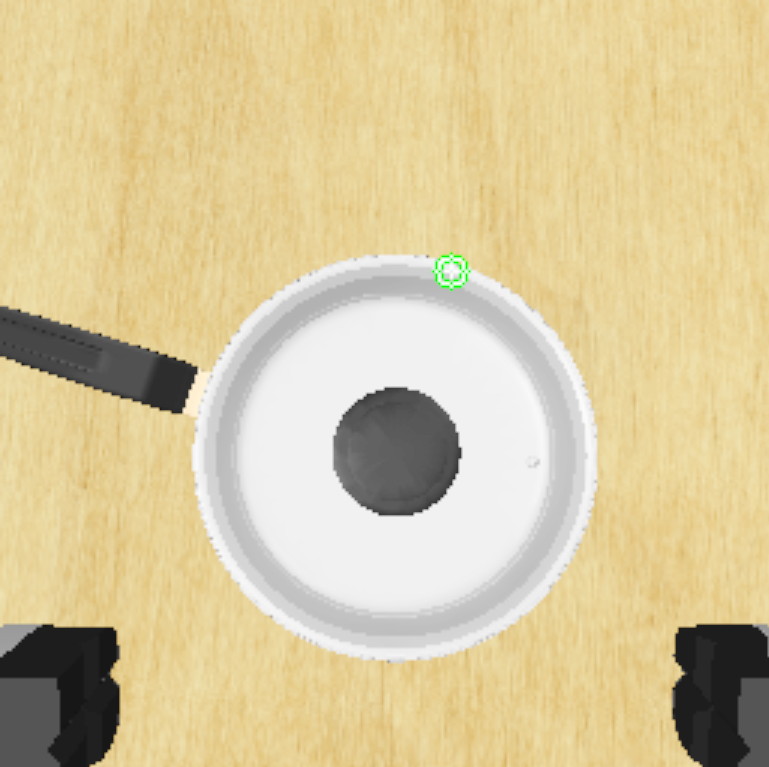}}
        \begin{minipage}[t]{\imagewidth}
            \centering
            \includegraphics[width=\imagewidth]{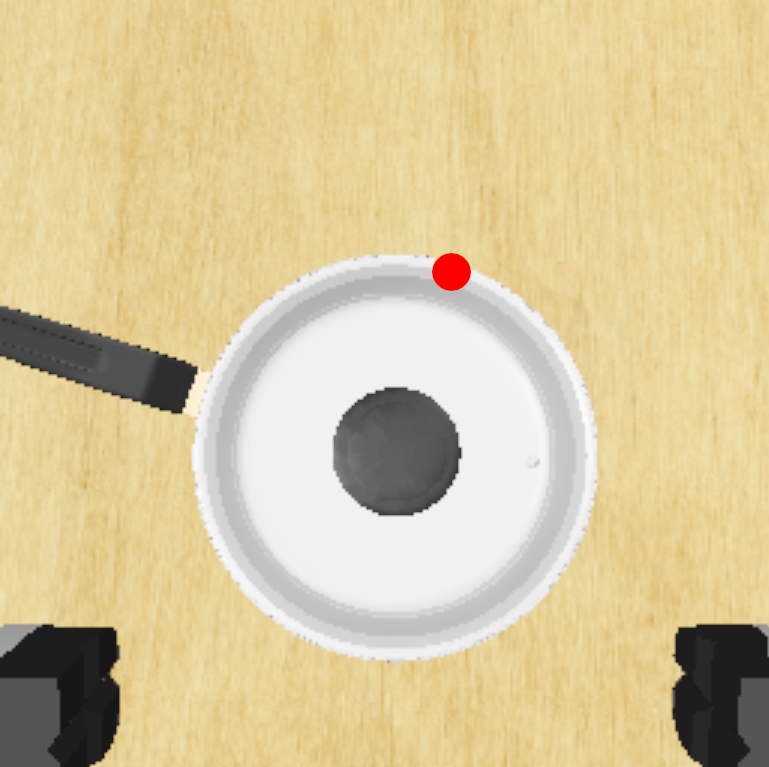}\\
            \footnotesize Reference
        \end{minipage}\hfil
        \begin{minipage}[t]{\imagewidth}
            \centering
            \includegraphics[width=\imagewidth]{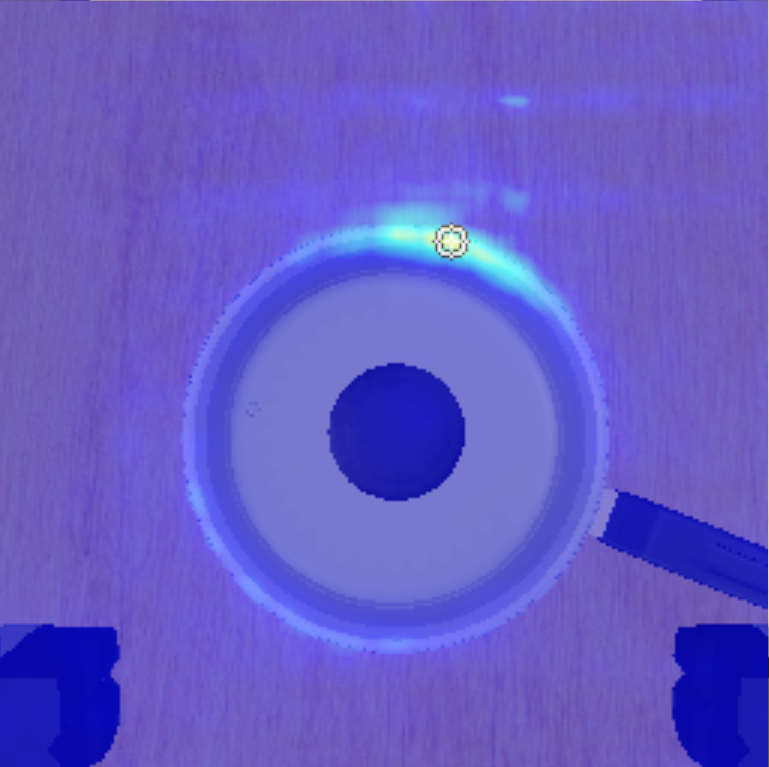}\\
            \footnotesize Base DON
        \end{minipage}\hfil
        \begin{minipage}[t]{\imagewidth}
            \centering
            \includegraphics[width=\imagewidth]{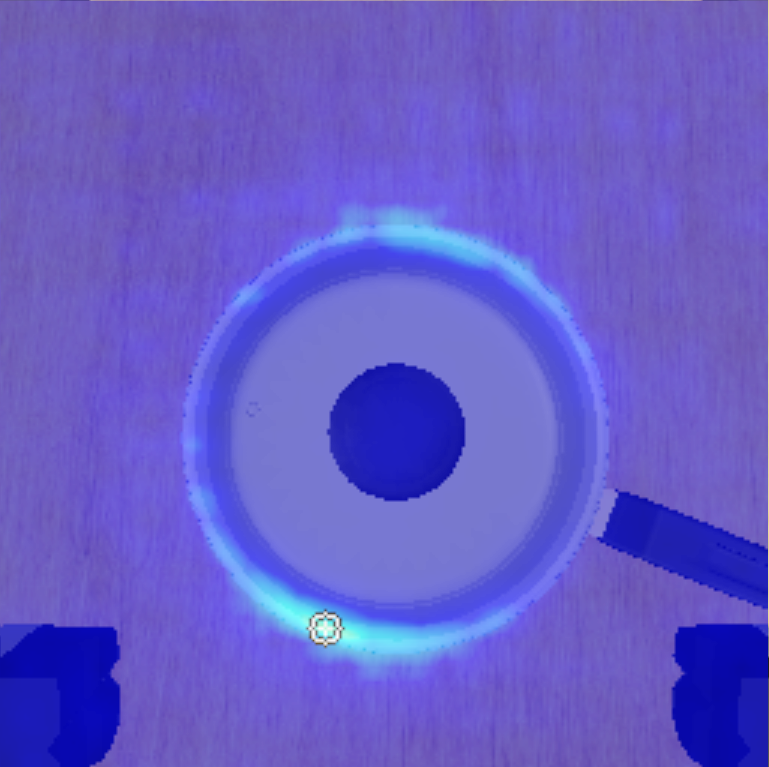}\\
            \footnotesize Ours: Rotation Equivariance
        \end{minipage}\hfil
        \begin{minipage}[t]{\imagewidth}
            \centering
            \includegraphics[width=\imagewidth]{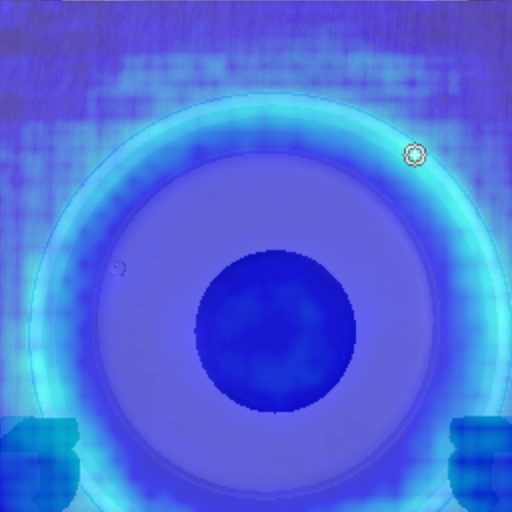}\\
            \footnotesize Ours: Multimodality
        \end{minipage}
        \caption{Rotation equivariance and multimodal correspondence distributions.
        The first image shows the reference image and reference location.
        The remaining images are overlaid with a heatmap indicating their correspondence likelihood. 
       In contrast to the base DON, our version effectively addresses both rotation equivariance and scale invariance, with multimodal hypotheses emerging when the \rebuttal{spatial} context is insufficient for unique correspondence.}\label{fig:lid_corr}
        \vspace{-0.3cm}
\end{figure}

\paragraph{\rebuttal{Rotation}}
Furthermore, many objects are partially symmetrical.
This makes rotation equivariance of the descriptor image an important property to generate consistent keypoints.
Similar to scale-invariance, achieving rotation equivariance requires higher semantic features and only emerges late in the training process.
Consequently, for partially symmetrical objects, the network needs to integrate information across the full image to resolve local symmetry, as shown in \figref{fig:lid_corr}.
Adding random rotations in training further helps the network generate more rotation equivariant descriptors.

\paragraph{\rebuttal{Occlusion}}
We find that larger descriptor dimensions and a deeper network enable training the encoder on more perspectives without loss in quality.
Thus, while previous work~\cite{florence2019self} uses the ResNet-34 model and descriptor size of 16, we use the ResNet-101 and descriptor size of 64.
To improve training on large descriptors, we normalize the descriptor distances by the square root of the descriptor dimension.
Moreover, during pretraining, we add aggressive crops of random size up to half of the dimension of the image.
We found that these enable the network to generalize better and to improve its robustness towards occlusions, especially when the size of the crops is randomized as well.
These adaptations enable us to train the DON to generate more semantically meaningful descriptors.
As we detail in \secref{sec:real_worl_exp}, our DON even shows zero-shot transfer to unseen object instances \rebuttal{and task environments}. 
Even more importantly, our DON consequently produces multimodal \rebuttal{hypothesis distributions} if the visual context is insufficient for unique localization, as shown in \figref{fig:lid_corr}.
This allows us to find likely correspondences across scales and then to resolve the multimodality of the hypothesis using context, as we detail in the subsequent section.

\subsubsection{Keypoint Inference}\label{sec:kp_gen}
\begin{figure}
    \centering
    \resizebox{!}{.33\linewidth}{\input{figures/histo1-histo.pgf}}\hspace{1em}%
    \raisebox{13pt}{\includegraphics[width=.27\linewidth]{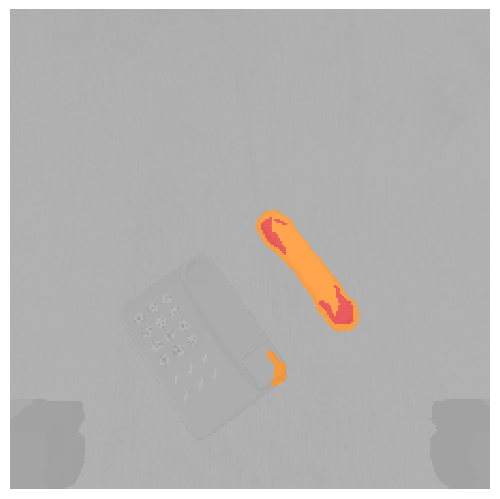}}
    \caption{Our correspondence likelihood model \(p(z) = e^{-4z}\) overlaid on a log-log histogram of an example distance map, next to a binned version of the map.
    \(M_{fg}\) and \(M_{bg}\) denote the margin parameters for foreground and background non-matches, respectively.
    As desired, the correspondence likelihood sharply declines for background pixels.
    }\label{fig:pf_meas_model}
    \vspace{-0.3cm}
\end{figure}

For policy learning, we sample one frame from the set of training trajectories \rebuttal{and feed it through the DON to select reference descriptors from.
In a GUI, we select the reference positions that we wish to track by clicking on them.}
The descriptors at these reference positions serve as the \emph{reference descriptors}.
We encode a camera observation by feeding it through the frozen encoder and computing the Euclidean distance map between each of the reference descriptors and the image embedding.
Subsequently, we employ a softmax function on the negative distance map to yield an activation map, interpreted as the probability of each pixel location corresponding to the reference position, \rebuttal{i.e.\ \(a_{I, r}=\sigma(\| e_\eta(I) - r\|)\) for an image \(I\) and reference descriptor \(r\)}.
To reduce the effect of background noise, we add a \emph{temperature} \(\alpha > 1\) to the softmax, as shown in \figref{fig:pf_meas_model}. 

For an unambiguous image that generates a unimodal correspondence map, we can use the expectation of this distribution \rebuttal{\(\mathbb{E}[a_{I, r}]\)} as the keypoint location \rebuttal{of \(r\) in \(I\)}.
By adding the associated depth measurement, we obtain a 3D keypoint.
In contrast, for an ambiguous image, it is first necessary to collapse the multimodality of the hypothesis distribution, as we discuss in the next section.

%% file: sections/scene_representation.tex
\subsection{Bayesian Scene Keypoints}\label{sec:bask}
Our goal is to integrate the keypoint localization hypotheses from successive camera observations to resolve ambiguities and reduce noise.
We propose an approach based on the Bayes filter which has \rebuttal{four} main advantages for our purpose.
First, it works with any number of observations, whereas \emph{learning} a sequence model would again require additional training data.
Second, it is easy to interpret and transparent in its assumptions, \rebuttal{which} renders it easy to debug, extend, and adapt to new situations.
For example, to integrate an additional modality such as LiDAR, all we need to formulate is a measurement model for it.
Third, in contrast to sequence models such as an LSTM~\cite{hochreiter1997long}, the Bayes filter keeps the spatial structure of the hypothesis space at all times, which constitutes a powerful inductive bias.
Finally, as no additional training is needed, representation learning and policy learning can be entirely decoupled.
Shortening the length of backpropagation paths drastically improves computational efficiency.
We verify these advantages in \secref{sec:supp_lstm} in the supplementary material.

We first summarize the basic form of the Bayes filter, before applying it to the scene representation by formulating the appropriate motion and measurement models.
We develop two Bayes filters: a simple discrete filter for single-camera setups and a more powerful particle filter that can represent unobserved hypotheses and is suited for multi-camera setups.

\subsubsection{Background: Bayes Filter}
For a time step \(t\), let \(x_t\in\mathbb{R}^d\) be the state of the environment at that time, \(u_t\) the action taken by the agent and \(z_t\) the measurement made.
For brevity, let \(z_{a:b}:=z_a,\ldots\rebuttal{,} z_b\).
To calculate the posterior \(P(x_t\mid z_{1:t}, u_{1:t})\) of the state given all past measurements and actions, we use Bayes' rule.
As per the usual convention, \(\eta\) denotes the normalization component of Bayes' rule, here \(\eta_t:=p(z_t\mid z_{1:t-1}, u_{1:t})^{-1}\). 
With the Markov assumption, we get:
\begin{align}\label{eq:bayes}
        p(x_t\mid z_{1:t}, u_{1:t}) &= \eta_t\cdot p(z_t\mid x_t) \cdot \int p(x_t\mid x_{t-1}, u_{t})\nonumber\\
    &\ \cdot p(x_{t-1}\mid z_{1:t-1}, u_{1:t-1})\ dx_{t-1},
\end{align}
where \(P(z_t\mid x_t)\) is referred to as the measurement model and \(P(x_t\mid x_{t-1}, u_{t})\) as the motion model.
However, the integration in \eqref{eq:bayes} is not always computationally feasible.
Instead, in many applications, approximations of the posterior have been successfully utilized.
These include the Kalman filters that can express Gaussian beliefs and the particle filter for arbitrary hypotheses.

\subsubsection{Application}
We propose two models.
For a single-camera setup, a discrete filter can be used by defining the camera's pixel space as the hypothesis space (\(d=2\)). 
Here, the main objective is to resolve the multimodality of the hypothesis distributions.
For a multi-camera setup, as well as to better handle occlusions and objects outside the current view frustum, we propose a particle filter.
Here, the hypothesis space is given by the world coordinate space (\(d=3\)).
Whereas in the discrete filter, each pixel in the camera's pixel space is a hypothesis, in the particle filter the hypothesis distribution is defined by a set of particles.
Each particle has an associated 3D location and weight that together express the filter's posterior.
In both filters, the location and weight of the hypotheses are updated using the motion and measurement model respectively.
After each update, we normalize the hypothesis likelihoods and return the weighted average of all hypotheses.

\begin{figure}[tb]
        \includegraphics[width=.245\linewidth]{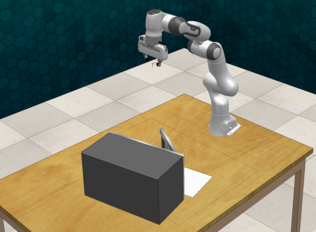}%
        \includegraphics[width=.245\linewidth]{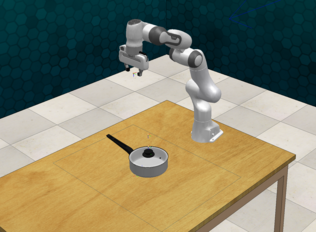}%
        \includegraphics[width=.25\linewidth]{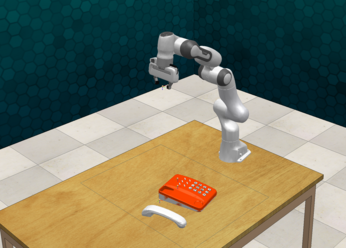}%
        \includegraphics[width=.25\linewidth]{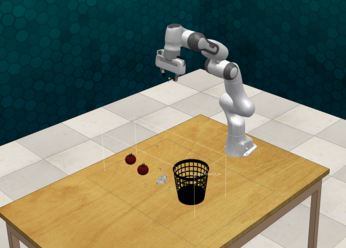}
        \includegraphics[scale=0.48]{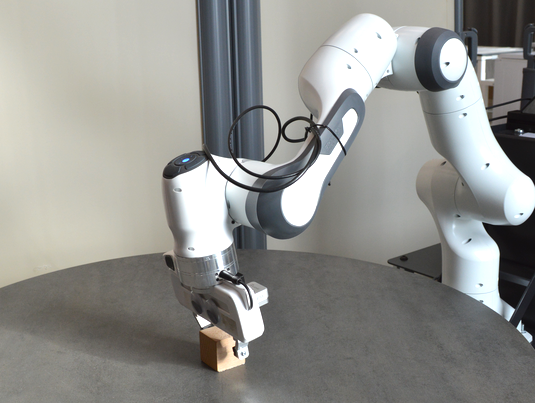}%
        \includegraphics[scale=0.48]{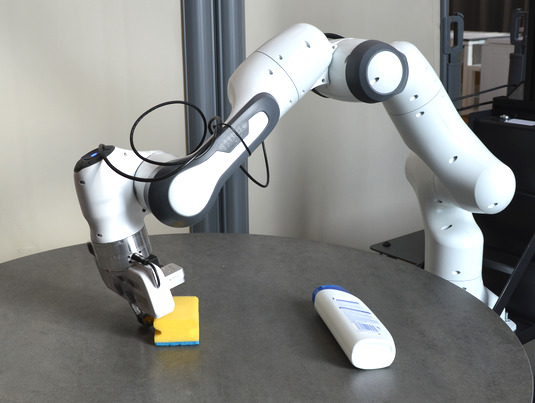}%
        \includegraphics[scale=0.48]{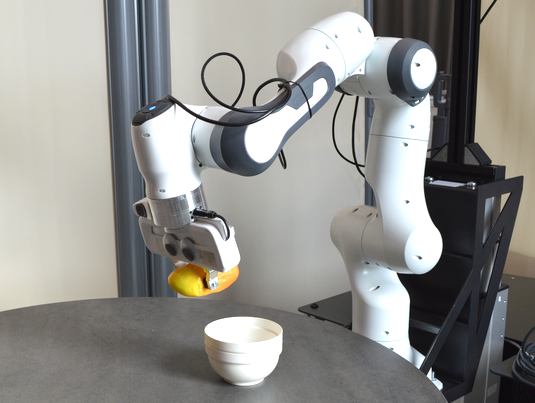}%
        \includegraphics[scale=0.48]{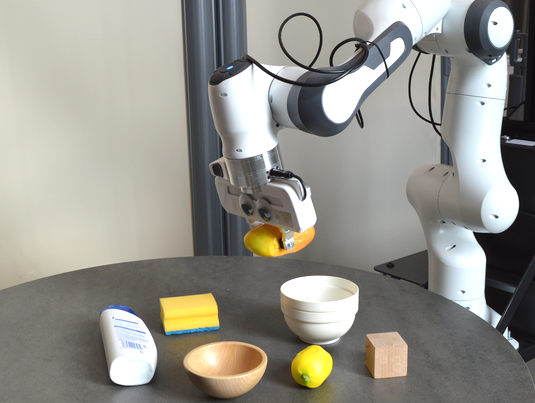}
        \caption{RLBench Tasks: \closemicrowave, \takelidoffsaucepan, \phoneonbase, \putrubbishinbin.
        Real-world tasks: \pickup, \pushtogoal, \pickandplace, and \pickandplace with clutter.}\label{fig:tasks}        \label{fig:real_tasks}
        \vspace{-0.3cm}
\end{figure}

\subsubsection{Motion Model}
For the discrete filter, the hypothesis space is the camera's pixel space.
Thus, at each time step, we correct for the movement of the camera by projecting the hypothesis distribution from the previous pixel space onto the current one.
For the particle filter which is anchored in the world frame, this is not necessary.
Meanwhile, in both filters, the movements of scene objects are modeled using a random walk with a fixed magnitude.
\begin{equation}
    \overline{x}_t = x_{t-1} + m_r,\quad m_r\sim\mathcal{N}(0, \sigma^2_r)
\end{equation}

Additionally, for the particle filter, we use the observation that the movements of scene objects are correlated with the movements of the robot gripper.
This allows us to reduce the required magnitude of the random walk, by stochastically applying the gripper motion \(G_t\) to each particle as
\begin{equation}
    \overline{x}_t = x_{t-1} + m_r + m_w\cdot G_t,\quad m_w\sim\mathcal{B}(p_w)
\end{equation}
where \(\mathcal{N}\) denotes a Gaussian and \(\mathcal{B}\) a Bernoulli distribution.
\(\sigma^2_r, p_w\) are hyperparameters.

\subsubsection{Measurement Model}
For the discrete filter, we use the correspondence model of the DON.
Recall that this is given by the softmax \(\sigma(z_t)\) over the current correspondence map as
\begin{equation}
    x_t = \overline{x}_t \odot \sigma(z_t).
\end{equation}
%

In contrast, for the particle filter our measurement model consists of three components: a correspondence likelihood model \(p_c\), a depth likelihood model \(p_d\), and an occlusion model \(p_o\).
As any particle might not be observed in an observation, we cannot use the softmax over the current observation as the correspondence model.
Instead, we use the underlying exponential function but omit the normalization component
\begin{equation}
    p_c(z_t\mid x_t) = e^{-\alpha\cdot z_t},
\end{equation}
where the temperature \(\alpha\) is a hyperparameter that allows us to tweak the measurement model for the expected value range, as shown in \figref{fig:pf_meas_model}.

\begin{figure*}[tb]
    \begin{minipage}[b]{0.64\textwidth}
        \begin{subfigure}[b]{0.135\linewidth}
            \includegraphics[width=\linewidth,]{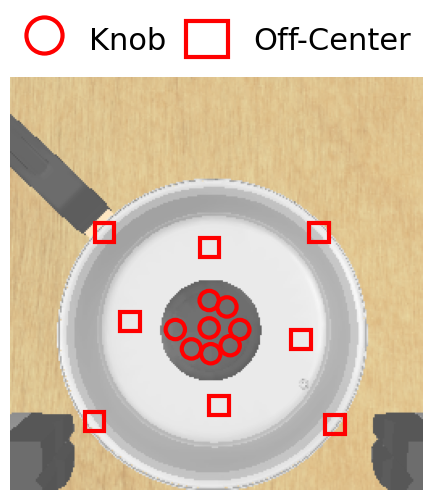}\\
            \vspace{5.25mm}
            \caption{Reference positions.}
        \end{subfigure}\label{subfig:ref_sel}\hfil
        \begin{subfigure}[b]{0.39\linewidth}
            \includegraphics[width=\linewidth]{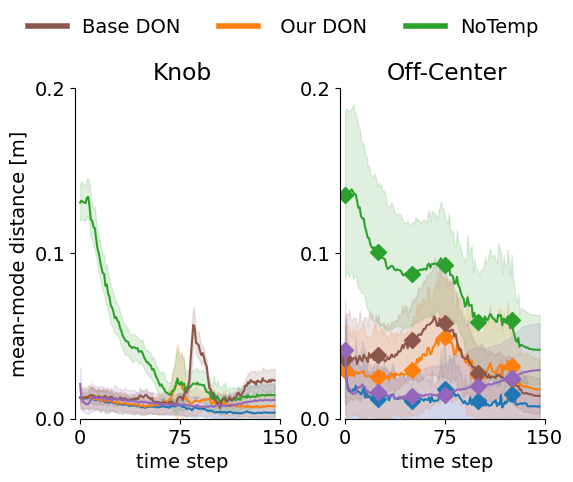}
            \caption{Multimodality (distance between mean and mode) of the activation map, aggregated.}\label{subfig:df_mm_dist}
        \end{subfigure}\hfil
        \begin{subfigure}[b]{0.39\linewidth}
            \includegraphics[width=\linewidth]{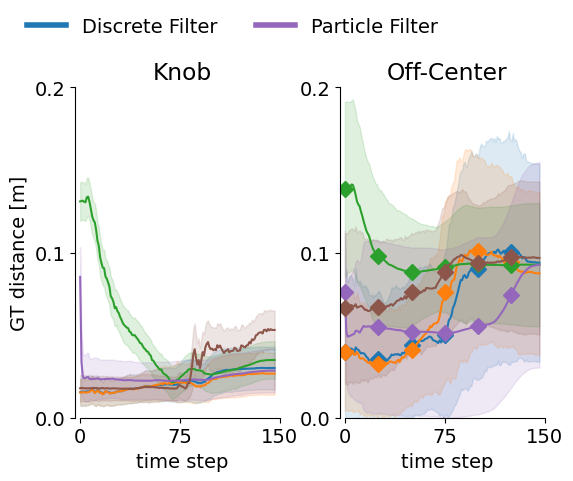}
            \caption{Distance to ground truth prediction of 3D-projected keypoints, aggregated.}\label{subfig:df_gt_dist}
        \end{subfigure}
        \par\bigskip
        \begin{subfigure}[t]{\linewidth}
            \centering
            \begin{overpic}[width=\textwidth, trim=70px 0 0 0, clip]{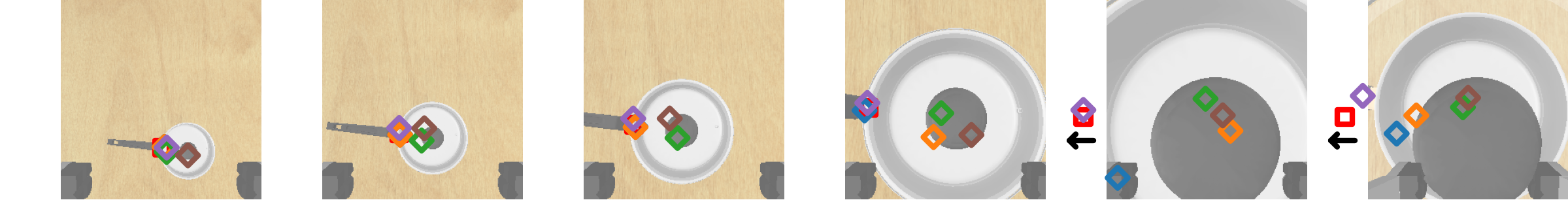}
             \put (33,36) {\footnotesize$t_0$}
             \put (86,36) {\footnotesize$t_{25}$}
             \put (143,36) {\footnotesize$t_{50}$}
             \put (199,36) {\footnotesize$t_{75}$}
             \put (251,36) {\footnotesize$t_{100}$}
             \put (307,36) {\footnotesize$t_{125}$}
            \end{overpic}
            \caption{Example trajectory of one keypoint. Red: ground-truth position.}\label{subfig:df_ex_traj}
        \end{subfigure}
        \caption{Correspondence quality for symmetrical objects.
        Thick lines and shaded areas indicate mean and standard deviation across trajectories and keypoints.
        As predicted, the unfiltered DONs produce a multimodal hypothesis for the off-center positions, when the spatial context is removed ($t_{75}$), leading to a steep increase in localization error.
        Both the discrete and particle filter resolve these multi-modalities.
        While the discrete filter, too, starts to fail when the ground truth position moves outside the field of view ($t_{100}$), the particle filter does not but only decreases in accuracy later due to the transparency of the lid ($t_{125}$).
        Both also show the effectiveness of using a temperature in the softmax for reducing the influence of background noise, as introduced in \secref{sec:kp_gen}.
        \subfiguresubref{subfig:df_ex_traj}  illustrates how the unmodified base DON (brown) fails to reliably find correspondences across the trajectory.
        }\label{fig:loc_so}
    \end{minipage}\hfill
    \begin{minipage}[b]{.32\textwidth}
        \begin{subfigure}[t]{\linewidth}
            \centering
            \includegraphics[width=.96\linewidth]{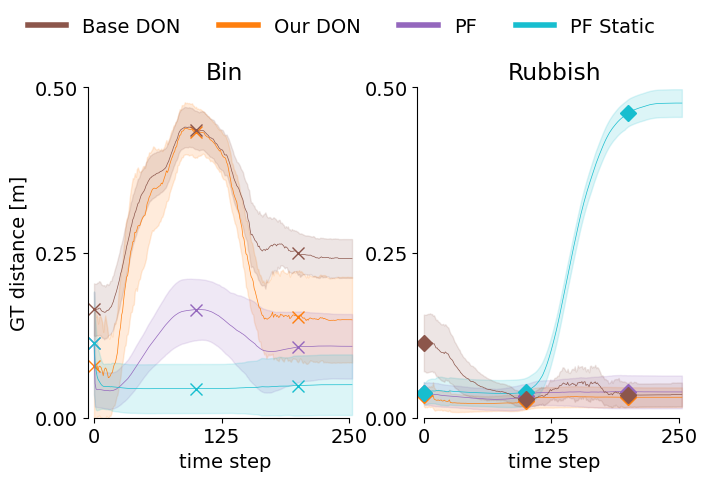}
            \caption{Distance to ground truth prediction of 3D-projected keypoints, aggregated across keypoints and trajectories.}\label{subfig:pf_gt_dist}
        \end{subfigure}\hfil
        \par\bigskip
        \begin{subfigure}[t]{\linewidth}
            \centering
            \begin{overpic}[height=43px]{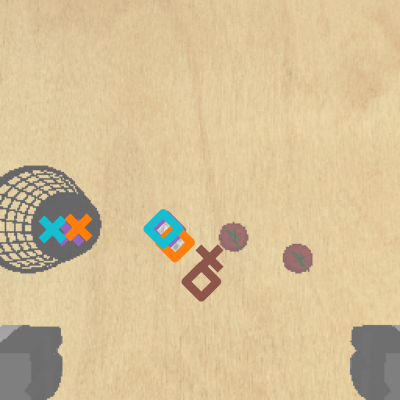}
             \put (34,36) {\footnotesize$t_0$}
            \end{overpic}
            \hfil
            \begin{overpic}[height=43px]{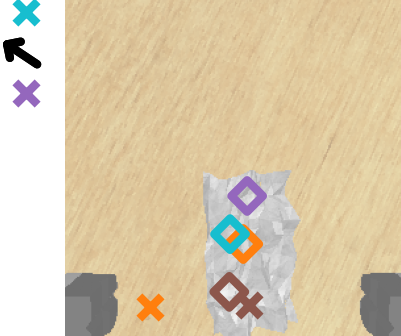}
             \put (36,36) {\footnotesize$t_{100}$}
            \end{overpic}
            \hfil
            \begin{overpic}[height=43px]{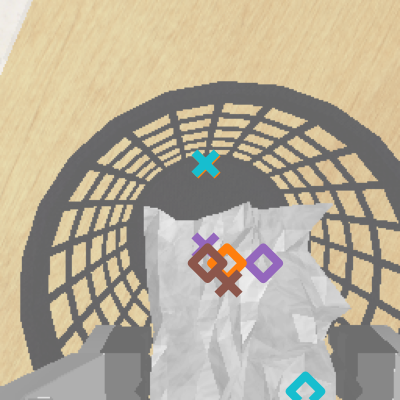}
             \put (28,36) {\footnotesize$t_{200}$}
            \end{overpic}
            \caption{Example trajectory of one keypoint per object.}\label{subfig:pf_ex_traj}
        \end{subfigure}
        \caption{
        Correspondence accuracy for unobserved and moving objects.
        While the unfiltered DON fails to accurately localize the rubbish bin ($\times$) outside the camera's fov ($t_{100}$), the particle filter does not.
        Injecting noise into the filter can reduce correspondence quality (purple), but the filter corrects when the bin comes back into view ($t_{200}$).
        In contrast, not doing so, disallows the model to track the moving piece of rubbish ($\diamond$, cyan).
        }\label{fig:loc_mo}
    \end{minipage}
    \vspace{-0.3cm}
\end{figure*}

To distinguish between particles with identical pixel coordinates but differing depths, we leverage an RGB-D camera.
We then factorize the measurement model into pixel-correspondence likelihood and depth likelihood.
To this end, we assume conditional independence of the depth measurement \(z_t^d\) and correspondence map \(z_t^v\), i.e.\ \(z_t^v\independent z_t^d \mid x_t\):
\begin{equation}
    p(z_t\mid x_t) = p(z_t^d, z_t^v\mid x_t) = p_d(z_t^d\mid x_t)\cdot p_c(z_t^v\mid x_t)
\end{equation}

The depth-likelihood \(p_d\) of a particle is then given by the noise model of the depth sensor.
We assume a zero-mean Gaussian, although more complex noise models can be used as well. 
So far, our measurement model is only suited for \emph{observed} particles, not for particles outside the view frustum and occluded particles.
To detect occlusions, we again use the depth model and compare the expected depth of a particle \(x_t^d\) with the measured depth \(z_t^d\) at the particle's position.
Additionally, we can leverage the assumption that an occlusion has only occurred if the depth difference \(z_t^d - x_t^d\) is larger than some margin \(\epsilon\).
This margin \(\epsilon \ge 0\) allows us to encode additional prior knowledge, leading to more robust occlusion detection.
For example, an occlusion caused by the robot or another object should lead to a depth difference of several centimeters.
In practice \(\epsilon = \SI{5}{\cm}\) works well, however, \(\epsilon=\SI{0}{cm}\) can be used to detect arbitrary occlusions as well.
Let \(p_o(z_t^d\mid x_t^d)\) denote the particle's occlusion likelihood.
Then:
\begin{align}
    p_o(z_t^d \mid x_t^d) &= 1 - p(z_t^d<x_t^d-\epsilon) \\
    &= 1 - \int_{-\infty}^{z_t^d} p_d(y \mid x_t^d-\epsilon)\ d y\label{eq:occ_model}.
\end{align}

In the case of occlusion, neither the descriptor \(z_t^v\) nor the depth measurement \(z_t^d\) is informative of the particle and we update the particle's likelihood using a constant value \(\tau\in (0,1)\).
The same holds true for particles outside the view frustum, which we detect by projecting their position onto the camera plane.
Let \(p_f(z_t):= \mathbf{1}_{\text{fov}}(z_t)\) indicate whether the particle lies \emph{inside} the view frustum.
Then, we combine all parts of our measurement model as

\begin{small}
\begin{align}
    p(z_t &\mid x_t) = p_o(z_t^d\mid x_t^d)\cdot p_f(z_t)\cdot \tau + (1 - p_f(z_t)) \cdot \tau \nonumber\\
    &+ (1-p_o(z_t^d\mid x_t^d))\cdot p_f(z_t) \cdot p_d(z_t^d\mid x_t)\cdot p_c(z_t^v\mid x_t)\label{eq:occ_distinction}.
\end{align}
\end{small}

\subsubsection{Initialization and Resampling}
We initialize the discrete filter using a uniform prior.
For the particle filter, we sample from the first correspondence likelihood map to ensure representativeness.
To avoid particle impoverishment, we use stratified resampling and resample as soon as the estimated fraction of effective particles \(\hat{N}_\mathit{eff} = \left(\sum_{i} w_i^2\right)^{-1}\) drops below a fixed threshold.
As some objects might not be visible in the first observation, we randomly resample particles from each observation with some small probability.
This technique allows the filter to quickly correct once the object becomes visible without impairing localization quality.

%% file: sections/experiments.tex
\section{Experimental Evaluations}
We perform experiments in simulation using RLBench~\cite{james2019rlbench}, a suite of manipulation tasks using everyday objects, and in the real world using the Franka Emika robot arm.
In both cases, the objects are placed randomly in the scene.
We evaluate on a challenging and representative subset of tasks, showing important \emph{visual} challenges in robotic manipulation.
In \closemicrowave, the policy is confronted with an articulated object, whereas in \takelidoffsaucepan there is high object symmetry and transparency of the lid.
\phoneonbase is a difficult multi-object task that requires careful alignment of the gripper, and
\putrubbishinbin introduces visual clutter.
The latter two tasks further introduce occlusions and the need to track multiple objects.
\rebuttal{We pre-train the dense correspondence for all tasks as described in \secref{sec:pretrain}.} 

\subsection{Correspondence Accuracy}
To verify that BASK is indeed able to resolve object symmetry and track objects across occlusions, we \rebuttal{record} 14 task demonstrations for \takelidoffsaucepan and 70 demonstrations for \putrubbishinbin using a wrist camera.
In \takelidoffsaucepan, we track 8 keypoints on the knob of the lid and 8 that are placed off-center.
In \putrubbishinbin we track 8 keypoints each on the static bin and the moving piece of rubbish.
The prediction accuracy is given by the Euclidean distance between the 3D-projected keypoints and the respective ground truth position.
\rebuttal{Results are aggregated across keypoints and trajectories.}
Furthermore, the distance between the expectation of the posterior density and its global mode (or $\argmax$) serves as an indicator for a multimodal \rebuttal{hypothesis distribution}.
We compare BASK and our improved DON to the original DON~\cite{florence2019self}, but for fairness use the same network architecture (ResNet-101), descriptor dimension (64), and softmax temperature, as well as normalizing the descriptor distance, as described in \secref{sec:inv}.
Furthermore, we compare against a variant of our DON without temperature. 

\begin{table*}[tb]
        \caption{Success rates of the learned policies in simulation for different representation learning methods. Mean, standard dev.\ across three training seeds.}\label{tab:success_rates_sim}
        \centering
        \begin{tabular}{l l c c c c c c}
            \toprule
            \textbf{Camera} & & \multicolumn{4}{c}{\textbf{Wrist-Only}} & \multicolumn{2}{c}{\textbf{+ Overhead}}\\
            \cmidrule(r){3-6}\cmidrule(l){7-8}
            & \diagbox[width=30mm, height=7.5mm]{Method}{Task} & Microwave & Lid & Phone & Rubbish & Phone & Rubbish\\
            \midrule
            \textbf{Baselines} & CNN & \( 0.72 \pm 0.09 \) & \( 0.93 \pm 0.04 \) & \( 0.62 \pm 0.03 \) & \( 0.39 \pm 0.26 \) & \(0.47 \pm 0.12\) & \(0.24 \pm 0.03\) \\
            & \(\beta\)-VAE~\cite{higgins2017beta} & \( 0.81 \pm 0.03 \) & \( 0.82 \pm 0.04 \) & \( 0.03 \pm 0.03 \) & \( 0.04 \pm 0.00 \) & \( 0.06 \pm 0.06 \) & \( 0.07 \pm 0.01 \) \\
            & Transporter~\cite{kulkarni2019unsupervised} & \( 0.75 \pm 0.04 \) & \( 0.98 \pm 0.04 \) & \( 0.54 \pm 0.08 \) & \( 0.43 \pm 0.04 \) & \( 0.67 \pm 0.02 \) & \( 0.46 \pm 0.12 \) \\
            & MONet~\cite{burgess2019monet} & \( 0.82 \pm 0.01 \) & \( 0.90 \pm 0.12 \) & \( 0.61 \pm 0.02 \) & \( 0.70 \pm 0.10 \) & \( 0.72 \pm 0.02 \) & \( 0.73 \pm 0.05 \) \\
            & Keypoints~\cite{florence2019self} & \( 0.72 \pm 0.13 \) & \( 0.35 \pm 0.14 \) & \( 0.04 \pm 0.05 \) & \( 0.04 \pm 0.03 \) & \( 0.10 \pm 0.02 \)& \( 0.07 \pm 0.02 \) \\
            \cmidrule{1-8}
            \textbf{Ours} & \emph{Keypoints} & \(\mathbf{ 0.93 \pm 0.02 }\) & \( 0.92 \pm 0.02 \) & \(\mathbf{0.78 \pm 0.02} \) & \( 0.00 \pm 0.00 \) & \(\mathbf{0.78 \pm 0.03} \) & \(\mathbf{0.92 \pm 0.02}\) \\
            & \emph{BASK: Discrete Filter} & \(\mathbf{ 0.93 \pm 0.02 }\) & \(\mathbf{ 0.99 \pm 0.01 }\) & \( 0.59 \pm 0.03 \) & \( 0.79 \pm 0.03 \) & \( 0.60 \pm 0.03 \) & \( 0.79 \pm 0.08 \) \\
            & \emph{BASK: Particle Filter} & \(\underline{0.92 \pm 0.02}\) &  \(\underline{0.98 \pm 0.01}\) & \(\underline{0.77 \pm 0.03}\) & \(\mathbf{0.92\pm 0.01}\) & \(\underline{0.77 \pm 0.03}\) & \(\mathbf{0.92\pm 0.01}\) \\
            \cmidrule{1-8}
            \textbf{Baseline} & Ground Truth \rebuttal{Keypoints} & \( 0.90 \pm 0.00 \) & \( 0.95 \pm 0.01 \) & \( 0.82 \pm 0.01 \) & \( 0.95 \pm 0.02 \) & \( 0.82 \pm 0.01 \) & \( 0.95 \pm 0.02 \) \\
            \bottomrule
        \end{tabular}
\end{table*}

\figref{fig:loc_so} presents the results on \takelidoffsaucepan and shows the emergence of a multimodal hypothesis for the DON as \rebuttal{scene} context is removed from the observation.
This is associated with a drop in accuracy which the discrete filter avoids.
Furthermore, the particle filter even stays accurate as the ground truth locations move outside the camera's field of view.
\figref{fig:loc_mo} shows the accuracy for \putrubbishinbin.
This plot highlights the particle filter's ability to recall the location of unobserved objects over long stretches of time, all the while tracking moving objects.
Without our improvements, the base DON fails to reliably track keypoints across scale differences and occlusions.
\figref{subfig:pf_ex_traj} in particular highlights how the base DON yields poor keypoint predictions in the presence of multiple objects, which our model effectively handles.

\subsection{Policy Learning in Simulation}
We train a 2-layer LSTM via behavioral cloning.
The action space is given by the change in the robot's end-effector pose and the observation space is given by the respective visual representation, concatenated with the robot's end-effector pose in the world frame.
For the particle filter, we add zero-mean Gaussian noise with \(\sigma = 0.02\) to the \rebuttal{predicted} keypoint locations \rebuttal{as we observe this to improve policy learning over using the low variance keypoint predictions directly.}
We provide the policy model with 14 demonstrations for the single-object tasks, using a wrist-mounted camera with \(256\times 256\) pixels.
For the multi-object tasks, we provide 140 demonstrations and use a stereo setup of overhead and a wrist camera with identical resolutions.
We train all the policies for 10,000 gradient steps on sub-trajectories consisting of 30 consecutive time steps~\cite{mandlekar2021matters} and on three random seeds.
We then evaluate all the policies in the respective task environments for 200 episodes.
We compare our method against a suite of other representation learning methods \rebuttal{introduced in \secref{sec:related_work}}, as well as a ground truth \rebuttal{keypoints} model.
Additionally, we \rebuttal{compare against an end-to-end trained 3-layer CNN with one downsampling layer and ELU activations as a baseline}.

\tabref{tab:success_rates_sim} shows the policy success rates.
The basic keypoints~\cite{florence2019self} fail on these tasks due to insufficient scale- and occlusion-invariance.
In contrast, our improved keypoints model outperforms the other representations across all the tasks and achieves a performance close to the ground truth model.
For the single-object tasks, the discrete filter, strictly improves over the bare keypoints model, even outperforming the ground truth model.
When only a few demonstrations are available, a small amount of noise in the representation leads to a more robust policy.
For example, on \takelidoffsaucepan, adding iid zero-mean Gaussian noise with \(\sigma = 0.05\) to the ground truth \rebuttal{keypoints} model's prediction leads to improved policy success of \(0.99\pm 0.01\).
On the multi-object tasks, data is more plentiful, with the ground-truth model matching the success rate of the human demonstrator.

Our improved keypoints model performs close to optimally as long as all relevant scene details are visible.
However, when learning from a wrist camera only, the model completely fails on \putrubbishinbin due to visual clutter.
The particle filter matches the keypoints model's performance in all cases (up to statistical certainty) and improves over it in a number of important cases.
Crucially, it is not affected by the clutter in \putrubbishinbin when learning from a wrist camera.
Moreover, \rebuttal{it is as effective
as the discrete filter} when confronted with the symmetrical lid and it is more sample efficient and robust towards observation dropouts, as shown in \secref{sec:ablations} in the supplementary material.

\subsection{Real Robot Policy Learning}\label{sec:real_worl_exp}
We perform real-world policy learning experiments on a Franka Emika robot arm with a wrist-mounted Intel Realsense D405 camera.
We compare the performance of our improved DON, BASK with the particle filter, a CNN baseline, and MONet~\cite{burgess2019monet} as the so far strongest competitor.
We design three tasks: \pickup, \pushtogoal, and \pickandplace, depicted in \figref{fig:real_tasks}.
We collect 35 demonstrations for \pickup, 50 demonstrations for \pushtogoal, and 100 demonstrations for \pickandplace.
All policies are trained for 15,000 gradient steps and evaluated for 50 episodes.

\begin{table}[tb]
        \caption{Success rates of the learned policies in real-world experiments.}\label{tab:success_rates_real}
        \centering
        \begin{tabular}{l l c c c}
            \toprule
            & & PickUp & PushToGoal & PickAndPlace\\
            \midrule
            \textbf{Baselines}  & CNN & 0.04 & 0.38 & 0.00 \\
            & MONet~\cite{burgess2019monet} & 0.08 & 0.44 & 0.00 \\
            \midrule
            \textbf{Ours}  & \emph{Keypoints} & 0.40 & 0.68 & 0.34 \\
            & \emph{BASK} & \(\mathbf{1.00}\) & \(\mathbf{0.88}\) & \(\mathbf{0.74}\) \\
            \bottomrule
        \end{tabular}
        \vspace{-0.3cm}
\end{table}

The challenges in real-world data clearly show in \tabref{tab:success_rates_real}.
Although MONet performs well in simulation, where idealized graphics allow it to easily segment objects by color, it is not effective in real-world experiments.
We provide qualitative insights in \secref{sec:monet_extra} in the supplementary material.
In contrast, our improved keypoints model performs strongly on the \pushtogoal task.
Nevertheless, BASK outperforms it substantially on multi-object tasks.
Furthermore, BASK stabilizes localization at close ranges, significantly improving policy success on the \pickup task as well.

As the supplementary videos show, BASK is robust towards visual clutter and manual intervention in the scene such as moving task objects.
Both these qualities emerge without additional training.
Even more, we observe zero-shot generalization of the policy to previously unseen object, \rebuttal{as reported in related work~\cite{florence2019self}, and to unseen task environments}.

%% file: sections/conclusion.tex
\section{Conclusions}
In this work, we introduce Bayesian Scene Keypoints (BASK) as a novel representation for robotic manipulation.
BASK overcomes the inherent limitations of current representation learning methods with respect to scale invariance and ambiguities, e.g.\ due to occlusions and a limited field of view.
It allows for efficient policy learning in multi-object scenes, especially when learning from wrist camera observations on a real robot.
Moreover, it facilitates robustness towards visual clutter and disturbances, as well as effectively generalizing to unseen objects.
Thus, it opens up a plethora of applications such as learning from wrist cameras, as well as flexible deployment in homes and in mobile manipulation.

Our Bayesian framework is agnostic towards the representation learning method itself as well as the policy learning approach.
Hence, it can be employed for alternative representations and pretraining schemes as well as novel policy learning methods.
Finally, using the particle spread as a certainty measure for the localization of a keypoint opens up further research directions toward Bayesian policies and active camera control for optimizing the localization certainty.

%% file: sections/supplementary.tex
\clearpage
\renewcommand{\baselinestretch}{1}
\setlength{\belowcaptionskip}{0pt}

\begin{strip}
\begin{center}
\vspace{-5ex}
\textbf{\LARGE \bf
The Treachery of Images: Bayesian Scene Keypoints\\\vspace{0.5ex}for Deep Policy Learning in Robotic Manipulation} \\
\vspace{3ex}

\Large{\bf- Supplementary Material -}\\
\vspace{0.4cm}
\normalsize{Jan Ole von Hartz$^{1}$, Eugenio Chisari$^{1}$, Tim Welschehold$^{1}$, Wolfram Burgard$^{2}$,\\ Joschka Boedecker$^{1}$ and Abhinav Valada$^{1}$}
\end{center}
\end{strip}

\setcounter{section}{0}
\setcounter{equation}{0}
\setcounter{figure}{0}
\setcounter{table}{0}
\setcounter{page}{1}
\makeatletter

\renewcommand{\thesection}{S.\arabic{section}}
\renewcommand{\thesubsection}{S.\arabic{subsection}}
\renewcommand{\thetable}{S.\arabic{table}}
\renewcommand{\thefigure}{S.\arabic{figure}}


\let\thefootnote\relax\footnote{$^{1}$Department of Computer Science, University of Freiburg, Germany.\\
$^{2}$Department of Engineering, University of Technology Nuremberg.
}%
\normalsize
In this supplementary material, we (i) illustrate why MONet~\cite{burgess2019monet} and other baselines perform poorly on real-world data, (ii) present additional ablation experiments on the data efficiency and robustness of our approach towards observations dropouts, (iii) give qualitative insights into why pose estimation methods fail in our experimental paradigm, (iv) show that LSTMs are unable to replace the Bayes filter in our setup, (v) list possible failure modes of our method, (vi) evaluate the generalization performance of our Dense Object Net, and (vii) summarize the multi-object mask generation.

\section{Real World Learning}\label{sec:monet_extra}
\begin{figure}
    \begin{subfigure}{\linewidth}
        \centering
        \includegraphics[width=\linewidth]{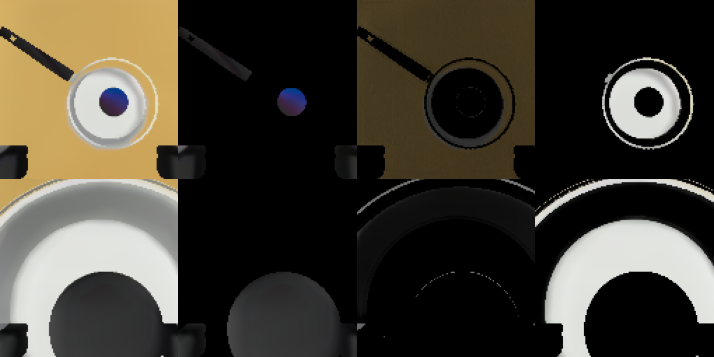}
        \includegraphics[width=\linewidth]{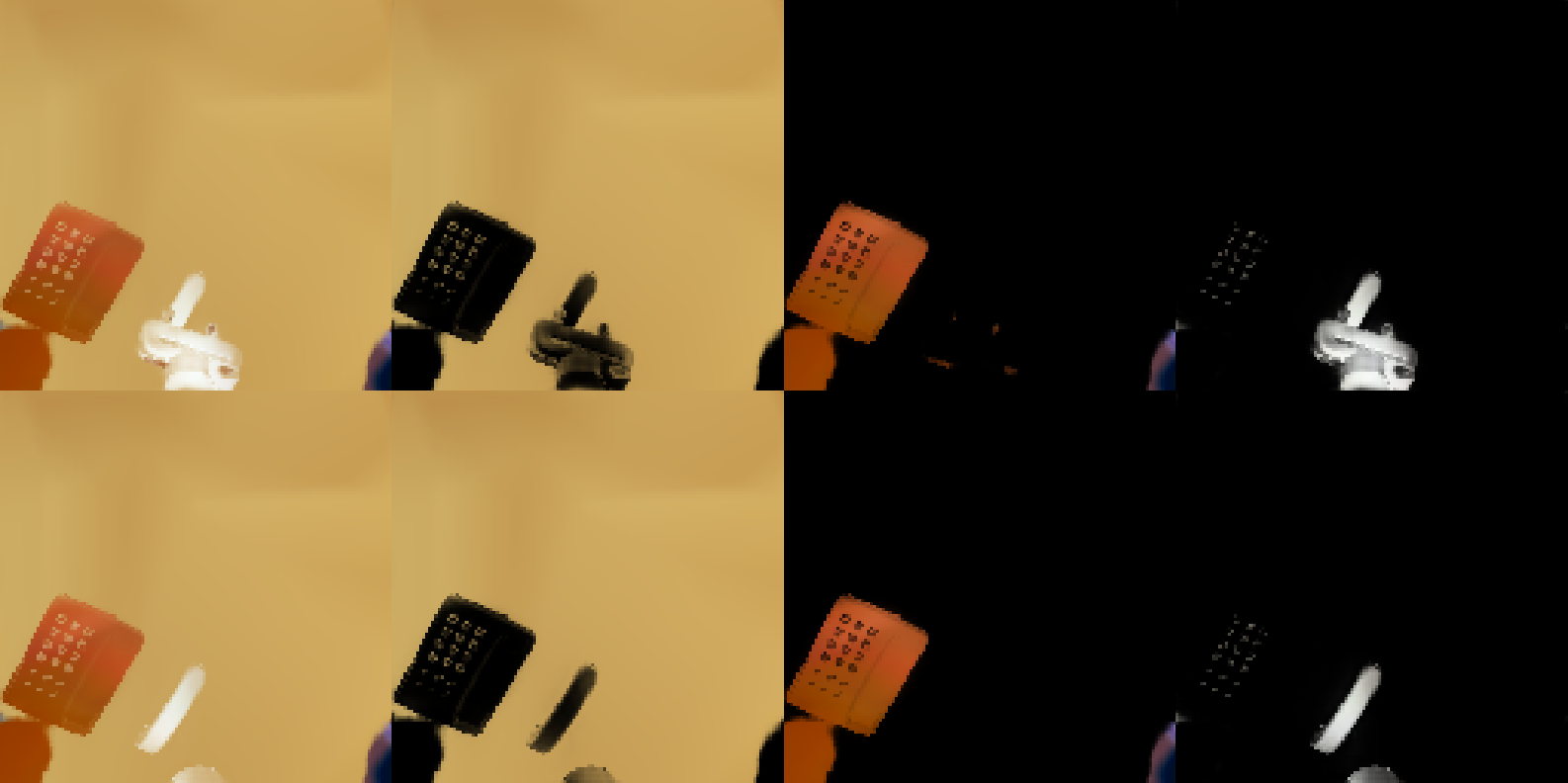}
        \caption{MONet in simulation}
        \label{fig:monet_sim}
    \end{subfigure}\\
    \begin{subfigure}{\linewidth}
        \centering
        \includegraphics[width=\linewidth]{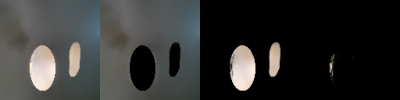}
        \includegraphics[width=\linewidth]{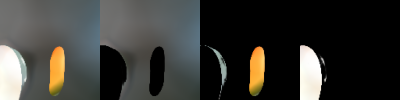}
        \caption{MONet in the real world}
        \label{fig:monet_real}
    \end{subfigure}
    \caption{MONet reconstruction results. The first column shows the full image reconstruction and the remaining columns show the individual slots.
    As well illustrated in \subfiguresubref{fig:monet_sim}, MONet tends to segment the image by color.
    It thereby even subdivides objects, as it can be seen with the saucepan.
    As the phone task illustrates, this strategy still works well in simulation due to the simplified graphics. Each object has a distinct color.
    However, as \subfiguresubref{fig:monet_real} shows, MONet fails on the more complex real-world data.
    While in the second row, it mostly succeeds in segmenting the objects, it fails to do so in the first row.
    Such inconsistent segmentation makes it difficult for the policy to implicitly infer object positions.
    }
\end{figure}

We present qualitative results to shed light on why MONet and other baselines perform poorly on real-world data compared to the simulation.
As briefly discussed in \secref{sec:related_work}, MONet partitions an image into several slots before auto-encoding the individual slots and slot masks.
The partitioning network and auto-encoders are trained jointly on an image reconstruction task.
Thus, it tends to partition the image by color and not necessarily by object borders, as well illustrated in \figref{fig:monet_sim}.
While it manages to differentiate between differently colored objects, e.g.\ the phone's base and receiver, it conflates objects with similar colors such as the receiver and robot arm.
At the same time, it further sub-partitions the lid into its differently colored parts.
While this strategy nevertheless works reasonably well in simulation, it fails on real-world data.
As \figref{fig:monet_real} illustrates, it fails to consistently partition the objects in the real-world scene, making it difficult for the downstream policy to learn the task at hand.

\begin{figure*}
    \includegraphics[width=.24\linewidth]{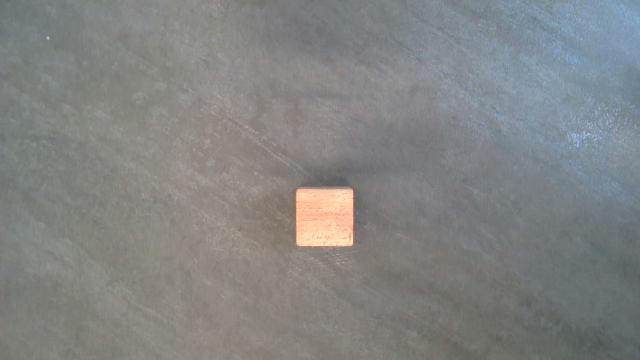}\hfil%
    \includegraphics[width=.24\linewidth]{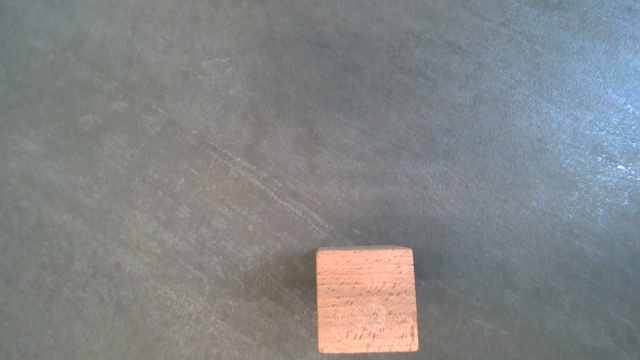}\hfil%
    \includegraphics[width=.24\linewidth]{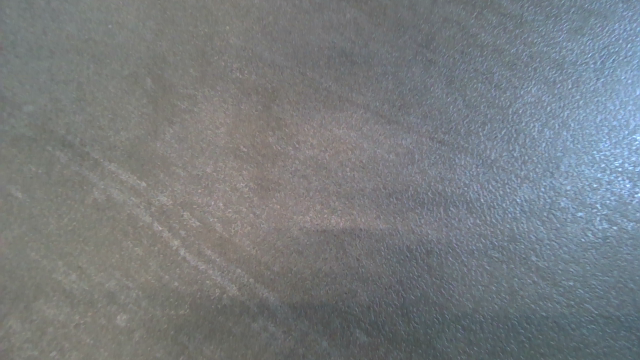}\hfil%
    \includegraphics[width=.24\linewidth]{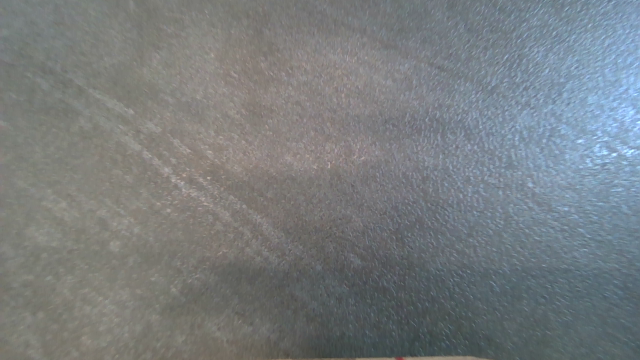}
    \caption{Wrist camera trajectory for the real-world \pickup task.
    Note how the task object is outside the field of view of the camera for significant parts of the trajectory.
    This includes the moment in time of the grasp attempt which happens at about the third frame shown above.
    Without the particle filter, the policy thus struggles to align the object and gripper.
    }\label{fig:real_tra}
\end{figure*}

Furthermore, comparing \figref{fig:real_tra} and \figref{subfig:df_ex_traj} illustrates why the CNN and keypoints model perform worse in the real world.
In the real-world task, object visibility along the trajectory is much reduced, making it harder for the policy to align the gripper and object.
BASK solves this problem as well.

\begin{figure*}
    \begin{subfigure}[t]{.5\linewidth}
        \resizebox{\textwidth}{!}{\input{figures/joint_demos.pgf}}
        \caption{Number of training demonstrations}\label{subfig:pol_demos}
    \end{subfigure}\hfil
    \begin{subfigure}[t]{.5\linewidth}
        \resizebox{\textwidth}{!}{\input{figures/joint_dropout.pgf}}
        \caption{Observation dropout }\label{subfig:pol_dropout}
    \end{subfigure}
    \caption[Policy success versus observation dropout and number of demonstrations]{
        Policy success versus observation dropout and the number of demonstrations.
        The thick lines indicate the mean success rates across three training seeds and the shaded area represents the standard error.
        In the dropout experiment, either camera observation is set to zero independently with identical probability.
        Note that while the dropout rate might seem rather high, we expect such independent dropouts to have a \emph{smaller} effect than the dropout of a sequence of continuous observations.}\label{fig:sim_ablation}
\end{figure*}
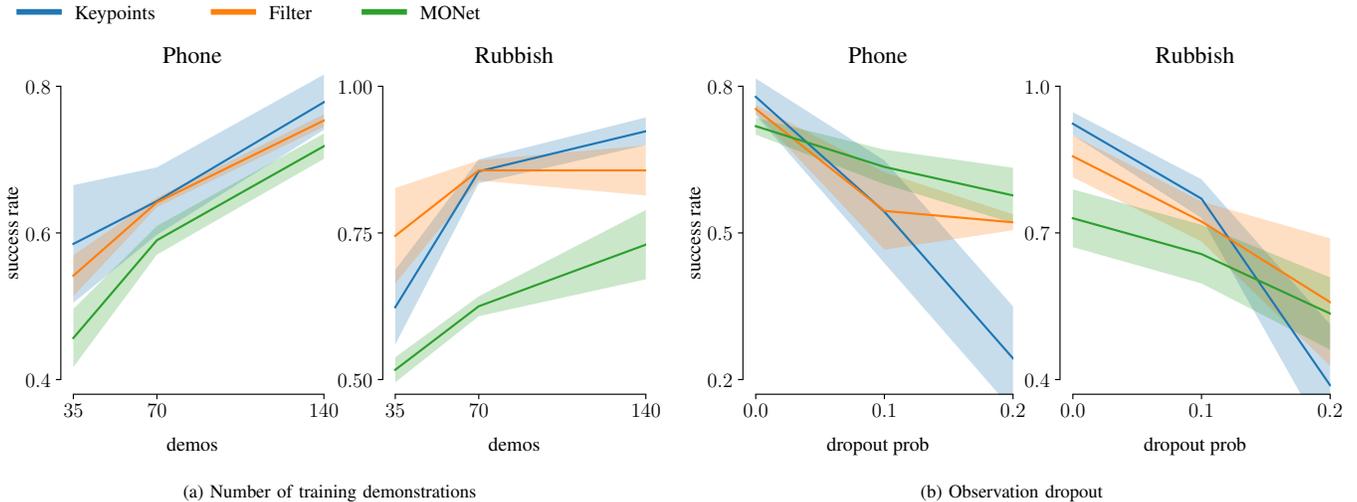

\begin{figure*}[h]
    \begin{subfigure}[t]{0.5\linewidth}
        \centering
        \includegraphics[width=\linewidth]{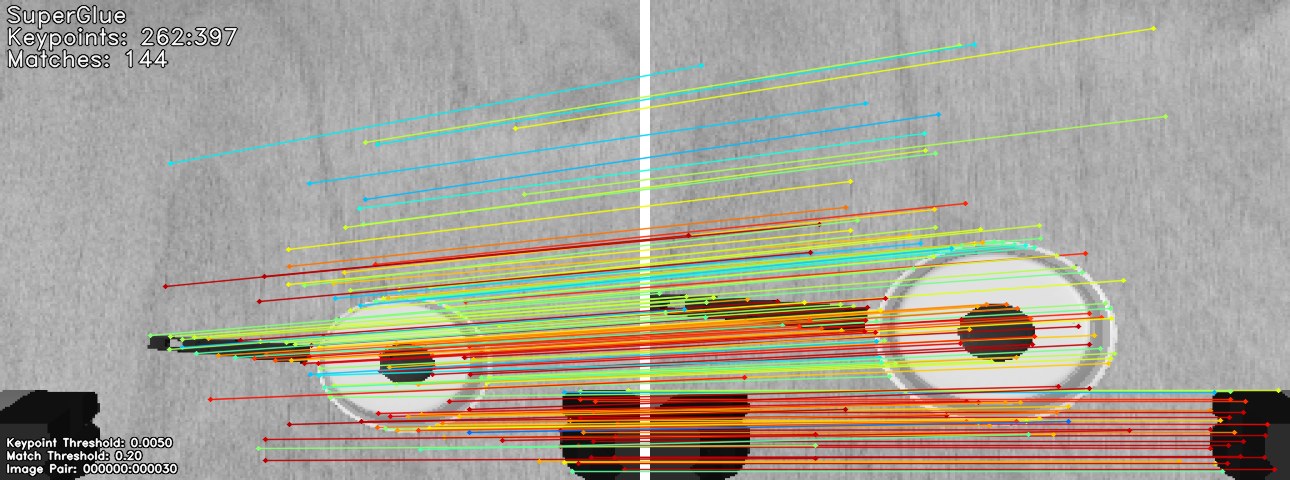}
        \includegraphics[width=\linewidth]{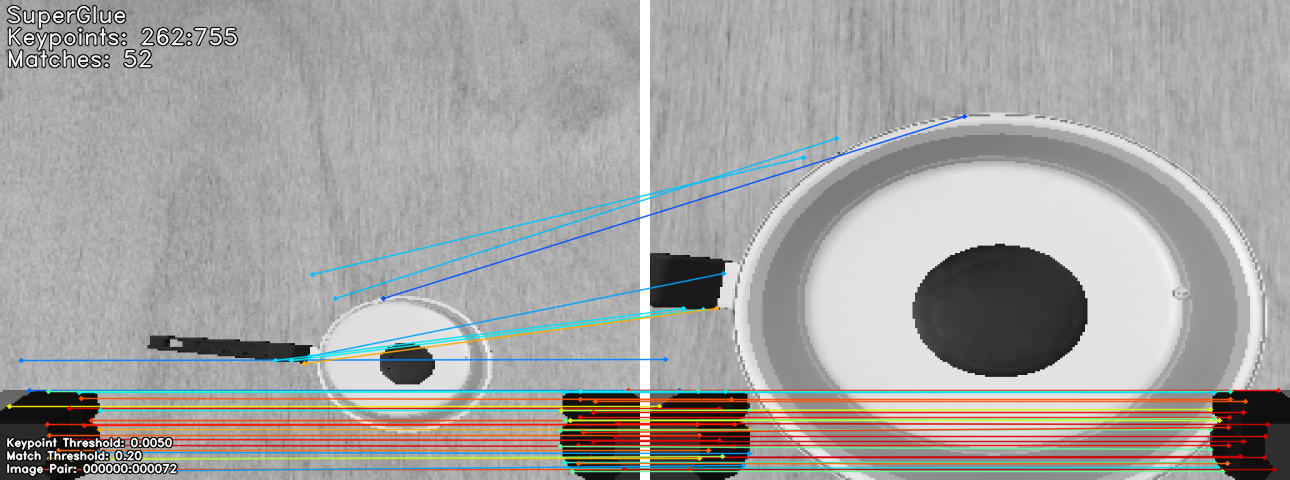}
        \includegraphics[width=\linewidth]{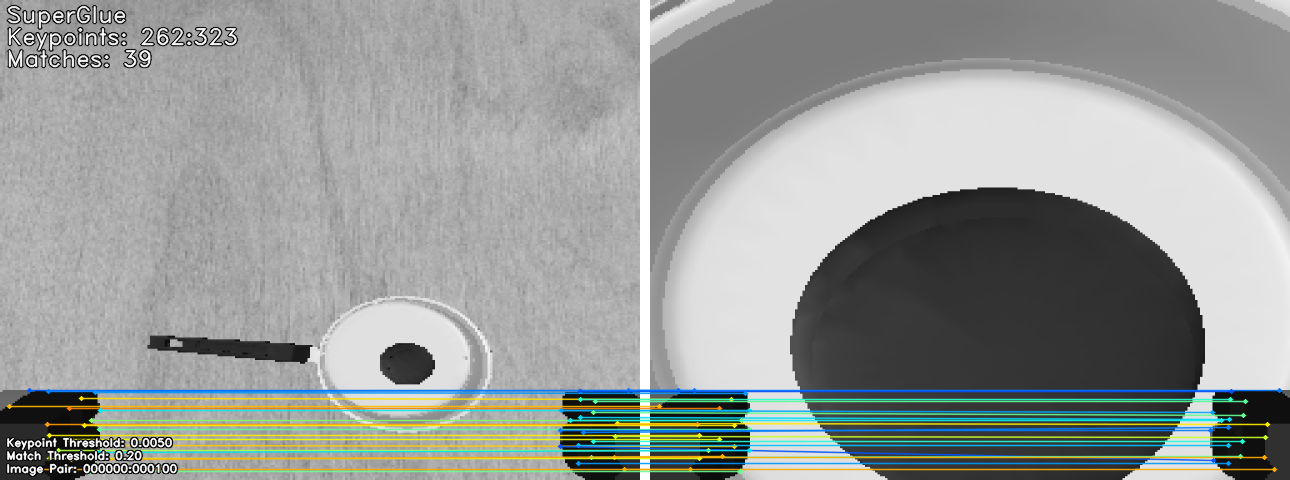} 
        \includegraphics[width=\linewidth]{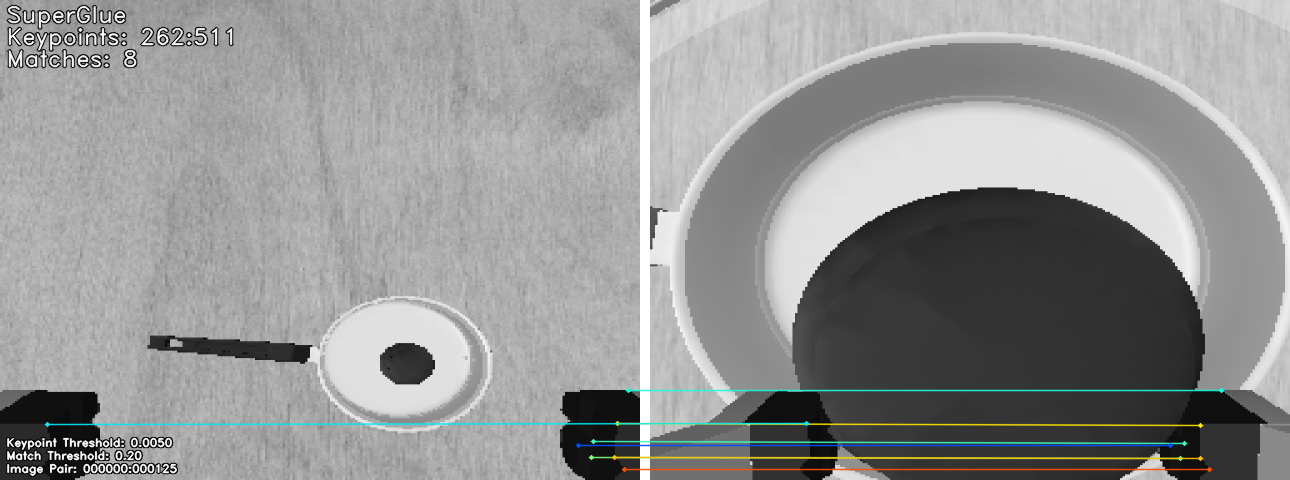} 
        \caption{SuperGlue feature correspondences on \takelidoffsaucepan.}
        \label{fig:sg_lid}
    \end{subfigure}
    \begin{subfigure}[t]{0.5\linewidth}
        \centering
        \includegraphics[width=\linewidth]{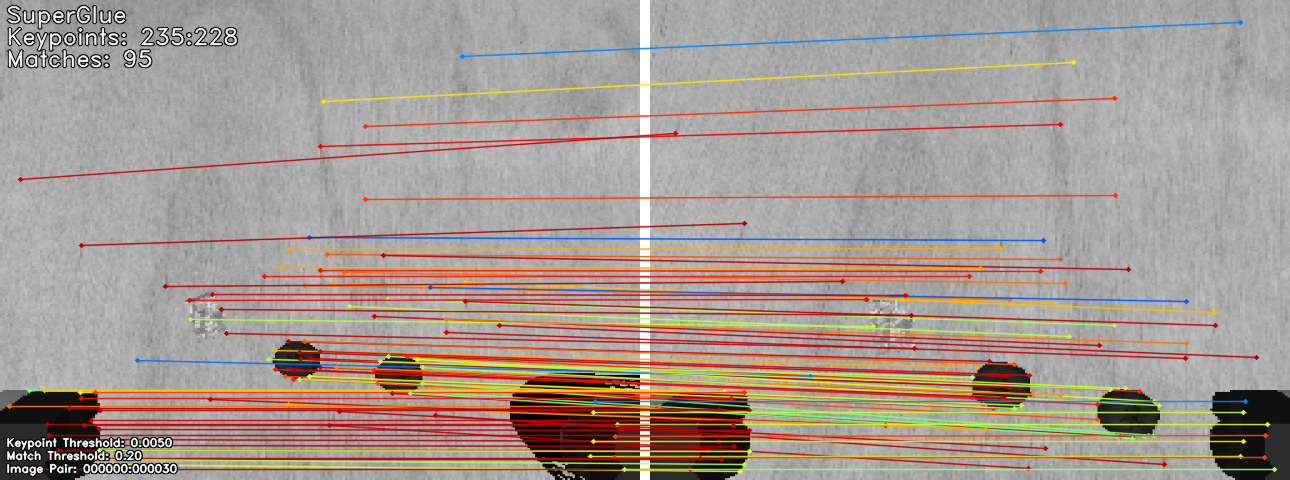}
        \includegraphics[width=\linewidth]{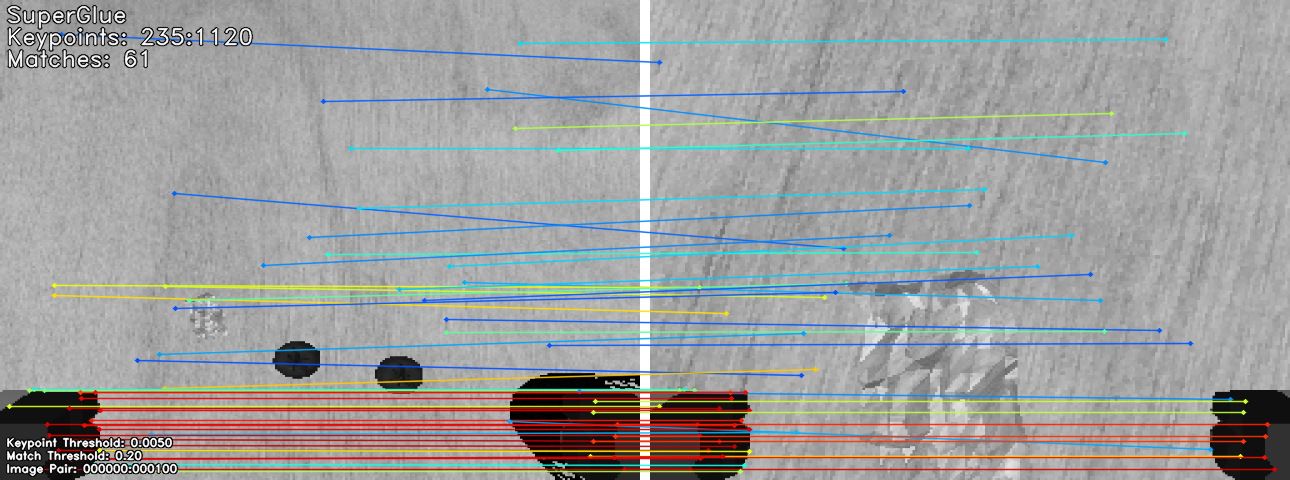}
        \includegraphics[width=\linewidth]{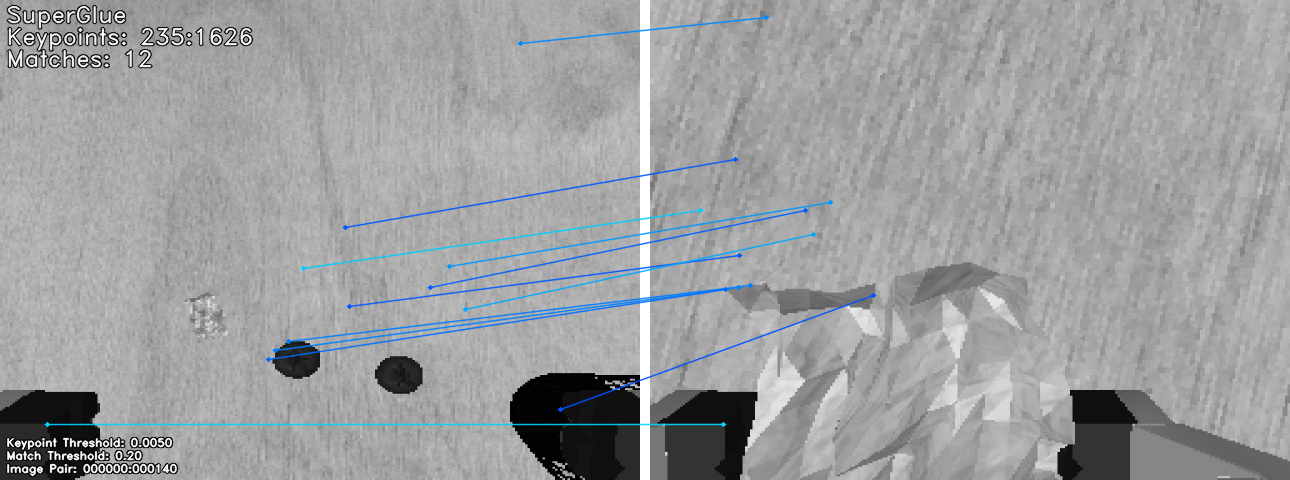} 
        \includegraphics[width=\linewidth]{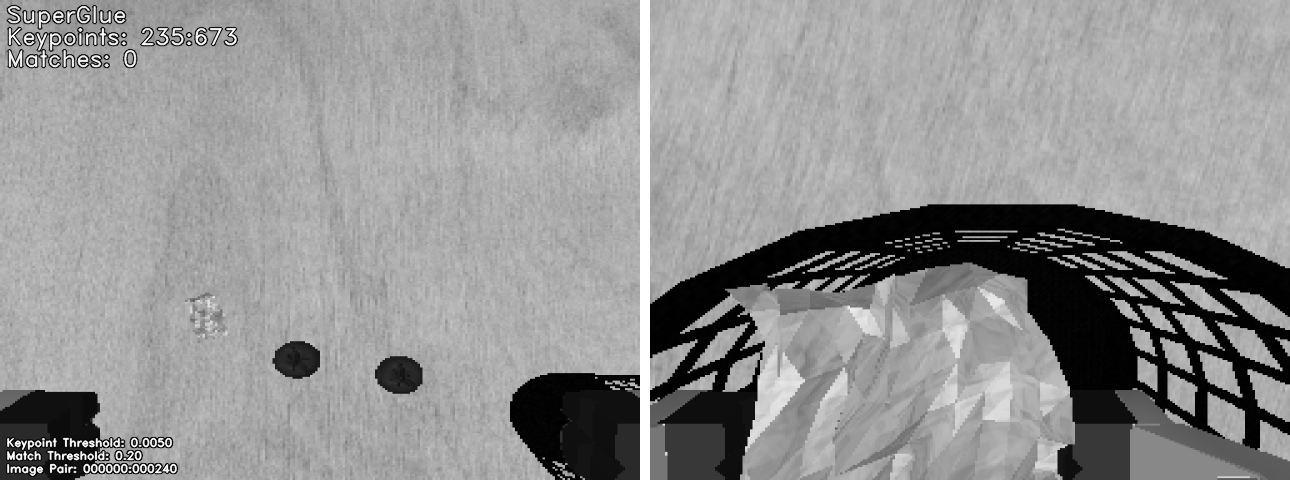} 
        \caption{SuperGlue feature correspondences on \putrubbishinbin.}
        \label{fig:sg_rubbish}
    \end{subfigure}
    \caption{SuperGlue~\cite{sarlin2020superglue} feature correspondence along task trajectories.
    Colored lines between image pairs indicate corresponding features, with red indicating high confidence in the prediction and blue low confidence.
    Images are shown in gray to improve the visibility of the matches.
    While SuperGlue finds correspondences between images that are visually similar (first row), it fails when scale-invariance is required.
    Already in the second row, it only finds high-confidence matches on the gripper, as well as low-confidence matches on the background, but almost no matches on the task objects.
    In the third row, it finds no matches between task objects at all.
    Interestingly, in the final row, it even fails to correspond points on the gripper.}\label{fig:superglue}
\end{figure*}

\section{Ablation Study}
\label{sec:ablations}

To investigate the sample efficiency of our approach as well as its robustness towards faulty observations, we perform experiments on \phoneonbase and \putrubbishinbin.
As we have already established that the pure keypoints model is less effective than BASK when learning from wrist camera only, we perform these experiments with wrist \emph{and} overhead camera.

For examining sample efficiency, we train policies following 
the same protocol as for our main experiments but with varying the number of human demonstrations.
\figref{subfig:pol_demos} indicates that BASK might be more sample efficient than the pure keypoints model.
Although we only observe an effect on \putrubbishinbin where the task dynamics are simpler but the \emph{visual} challenges are greater.
On \phoneonbase the task dynamics are more difficult, as precise rotations of the gripper are required but the visual challenges are smaller as there is no visual clutter.

For examining robustness, we randomly dropout observations, i.e.\ set them to zero, with varying probability. In line with our expectations, \figref{subfig:pol_dropout} suggests that BASK is more robust towards observation dropouts on both tasks. MONet seems to be robust towards observation dropouts but less sample efficient than BASK.

\section{Pose Estimation Methods}\label{sec:pose_est}
As we briefly discuss in \secref{sec:related_work}, pose estimation approaches have severe downsides compared to our method.
Most pose estimation methods rely on a ground truth 3D model, whereas none of the other methods we compare against has access to a ground truth model.
The reliance on a 3D model also limits pose estimators to specific object instances.
Moreover, they are not applicable to articulated or deformable objects.
In contrast, DONs can be flexibly applied to new objects, articulated and deformable objects, and even generalize between object instances.
Furthermore, pose estimation usually does not work well with occlusions or a limited field of view, whereas we study learning from wrist-camera observations where both these challenges arise.
While Gen6D~\cite{liu2022gen6d} and OnePose\cite{sun2022onepose}, two novel approaches, solve the reliance on a 3D model, they still require a pre-recorded object scan for inference and suffer from the remaining problems.

Finally and crucially, these pose estimators \emph{also} rely on corresponding visual features under-the-hood.
For instance, OnePose uses SuperGlue\cite{sarlin2020superglue}, which we found not to exhibit the strong scale invariance needed for our wrist camera setup.
This is shown in \figref{fig:superglue} for \takelidoffsaucepan and \putrubbishinbin.
SuperGlue only manages to correspond features for images that are visually similar and fails to correspond along a task trajectory.
Without the necessary correspondence of features, the subsequent pose estimation cannot proceed. Hence, adapting, for example, OnePose to make it work in our scenario would require both to replace the used features with a scale-invariant feature encoder, such as our improved DON, as well as adding a Bayes filter to stabilize predictions when objects are out of view.
Thus, we would need to alter the pose estimator to such a degree that it becomes almost indistinguishable from our approach.

\section{LSTMs as Sequence Models}\label{sec:supp_lstm}
In \secref{sec:bask}, we have argued that a Bayes filter is preferable over training an LSTM for integrating keypoint locations over time.
To verify these claims, we trained an additional one-layer LSTM in between the DON and the policy model as a replacement for the Bayes filter on \takelidoffsaucepan.
To be applicable to real-world tasks where the only supervision signal is the task demonstrations, this sequence model was trained end-to-end.
In line with our hypothesis, the predicted keypoint locations are vastly inaccurate, with a ground-truth distance of 
\(\SI{0.9}{\m}\pm\SI{0.1}{\m}\).
This also leads to the reduced success of the downstream policy of \(56\%\) compared to \(98\%\) of BASK.

We also find that decoupling representation learning and policy learning improves computational efficiency by shortening backpropagation paths.
In fact, by pre-encoding each observation, we observe a speedup of one order of magnitude during policy learning.

\section{Failure Cases}
\label{sec:failure_cases}
Failure cases of our method can be differentiated between correspondence (DON) failure and filter (BASK) failure.
As our supplementary video and \secref{sec:don_gen} shown,  our DON generalizes well, however, as with any learning-based approach, generalization is not unlimited.
For example, very drastic illumination changes and reflections pose a challenge to virtually all visual methods, including DONs.
Similarly, current depth cameras, which we use for 3D information, struggle with reflective surfaces.

\begin{table*}[tb]
        \caption{Generalization performance measured by the normalized Euclidean pixel distance (mean and std.).}\label{tab:genexp_results}
        \centering
        \begin{tabular}{l c c c c c}
            \toprule
            & Training & In-domain & Object & Environment & Both \\
            \midrule
            \textbf{Cups}  & \(0.055 \pm 0.042\) & \(0.055 \pm 0.044\) & \(0.058 \pm 0.044\) & \(0.054 \pm 0.041\) & \(0.060 \pm 0.057\) \\
            \textbf{Cans}  & \(0.073 \pm 0.055\) & \(0.072 \pm 0.058\) & \(0.088 \pm 0.060\) & \(0.073 \pm 0.052\) & \(0.086 \pm 0.059\) \\
            \textbf{Citrus} & \(0.051 \pm 0.034\) & \(0.052 \pm 0.036\) & \(0.069 \pm 0.037\) & \(0.054 \pm 0.037\) & \(0.067 \pm 0.032\) \\
            \bottomrule
        \end{tabular}
        \vspace{-0.3cm}
\end{table*}

BASK itself makes few assumptions, making it also fairly robust.
However, it can only incorporate the information that is actually available.
Thus, for objects that are occluded until the end of the trajectory, BASK can never update its localization estimate.
As our supplementary video further shows, the spatial consistency enforced by BASK allows to discriminate even between identical looking objects (\rebuttal{Ref.\ suppl.\ video}).
Though, there is a trade-off in the hyperparameter settings between enforcing temporal consistency and incorporating new information.
For example, for optimal performance, the magnitude of the motion model should to be set to roughly match the expected object speeds.

\section{Generalization Study}
\label{sec:don_gen}

In our supplementary video we have shown zero-shot transfer of our DON to unseen objects and environments.
To quantify the generalization performance of our Dense Object Net, we perform additional experiments using the ManiSkill2 benchmark~\citesuppl{gu2023maniskill2}, YCB object dataset~\citesuppl{calli2017yale} and a set of four unseen environments~\citesuppl{sketchfab1, sketchfab2, sketchfab3, sketchfab4}.

\begin{figure}[tb]
    \centering
    \includegraphics[width=.25\linewidth]{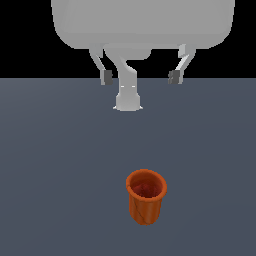}%
    \includegraphics[width=.25\linewidth]{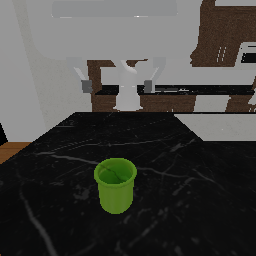}%
    \includegraphics[width=.25\linewidth]{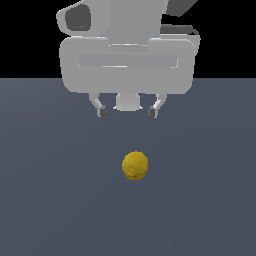}%
    \includegraphics[width=.25\linewidth]{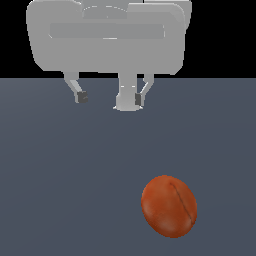}
    \caption{Example scenes from our DON generalization experiment. We evaluate the generalization between environments (e.g.\ blue background to kitchen), object instances (e.g.\ lemon to orange) and both.}\label{fig:genexp}
\end{figure}

We select three object sets from the YCB dataset: cups; cans; citrus (lemon and orange).
For each object set, we then collect 20 observations of the first instance of that object set on a neutral blue backdrop using the base camera, and train a DON on them.
Afterwards, for of all objects from the set, and in all environments, we collect 50 demonstrations each for evaluation.
We randomly sample 16 keypoints from a reference observation and estimate the prediction error of the DON by comparing its prediction to that of our ground truth keypoints model.
Results for environment generalization are aggregated across all four test environments.
We further report the error for an in-domain test set that was collected in the same way the training set was.
\figref{fig:genexp} illustrates the collected data.

As \tabref{tab:genexp_results} shows, both the in-domain error and environemnt-generalization error are comparable to the training set error.
The object generalization and joint object-environment generalization error are slightly larger, but still firmly within the margin of error of the estimates.
This indicates that the DON generalizes well across object instance and environments.
Instead, the biggest challenge seems to be the precise correspondence of visually generic features, such as points on the side of a monochrome cup.

\section{Multi-Object Mask Generation}
\label{sec:momg}

\begin{figure}[tb]
    \centering
    \includegraphics[width=\linewidth]{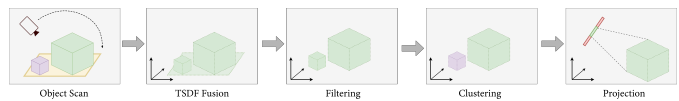}
    \caption{Multi-object mask generation.
    As in the single-object case, we reconstruct the 3D scene using volumetric reconstruction and filter out background vertices. The point cloud is then split into individual objects using density-based clustering, and the individual meshes are projected back onto the camera.}
    \label{fig:momg}
\end{figure}

As discussed in \secref{sec:mo} we train directly on static scans of multi-object scenes.
In \figref{fig:momg} we have again summarized the process of multi-object mask generation.

\footnotesize
\bibliographystylesuppl{IEEEtran}
\bibliographysuppl{references_appendix.bib}

%% file: figures/joint_demos.pgf
\begingroup%
\makeatletter%
\begin{pgfpicture}%
\pgfpathrectangle{\pgfpointorigin}{\pgfqpoint{6.687636in}{4.721612in}}%
\pgfusepath{use as bounding box, clip}%
\begin{pgfscope}%
\pgfsetbuttcap%
\pgfsetmiterjoin%
\pgfsetlinewidth{0.000000pt}%
\definecolor{currentstroke}{rgb}{0.000000,0.000000,0.000000}%
\pgfsetstrokecolor{currentstroke}%
\pgfsetstrokeopacity{0.000000}%
\pgfsetdash{}{0pt}%
\pgfpathmoveto{\pgfqpoint{0.000000in}{0.000000in}}%
\pgfpathlineto{\pgfqpoint{6.687636in}{0.000000in}}%
\pgfpathlineto{\pgfqpoint{6.687636in}{4.721612in}}%
\pgfpathlineto{\pgfqpoint{0.000000in}{4.721612in}}%
\pgfpathlineto{\pgfqpoint{0.000000in}{0.000000in}}%
\pgfpathclose%
\pgfusepath{}%
\end{pgfscope}%
\begin{pgfscope}%
\pgfsetbuttcap%
\pgfsetmiterjoin%
\pgfsetlinewidth{0.000000pt}%
\definecolor{currentstroke}{rgb}{0.000000,0.000000,0.000000}%
\pgfsetstrokecolor{currentstroke}%
\pgfsetstrokeopacity{0.000000}%
\pgfsetdash{}{0pt}%
\pgfpathmoveto{\pgfqpoint{0.705073in}{0.712467in}}%
\pgfpathlineto{\pgfqpoint{3.284570in}{0.712467in}}%
\pgfpathlineto{\pgfqpoint{3.284570in}{3.878179in}}%
\pgfpathlineto{\pgfqpoint{0.705073in}{3.878179in}}%
\pgfpathlineto{\pgfqpoint{0.705073in}{0.712467in}}%
\pgfpathclose%
\pgfusepath{}%
\end{pgfscope}%
\begin{pgfscope}%
\pgfpathrectangle{\pgfqpoint{0.705073in}{0.712467in}}{\pgfqpoint{2.579497in}{3.165712in}}%
\pgfusepath{clip}%
\pgfsetbuttcap%
\pgfsetroundjoin%
\definecolor{currentfill}{rgb}{0.121569,0.466667,0.705882}%
\pgfsetfillcolor{currentfill}%
\pgfsetfillopacity{0.250000}%
\pgfsetlinewidth{0.000000pt}%
\definecolor{currentstroke}{rgb}{0.000000,0.000000,0.000000}%
\pgfsetstrokecolor{currentstroke}%
\pgfsetdash{}{0pt}%
\pgfpathmoveto{\pgfqpoint{0.827906in}{2.764542in}}%
\pgfpathlineto{\pgfqpoint{0.827906in}{1.610260in}}%
\pgfpathlineto{\pgfqpoint{1.646794in}{2.275547in}}%
\pgfpathlineto{\pgfqpoint{3.284570in}{3.307346in}}%
\pgfpathlineto{\pgfqpoint{3.284570in}{3.849445in}}%
\pgfpathlineto{\pgfqpoint{3.284570in}{3.849445in}}%
\pgfpathlineto{\pgfqpoint{1.646794in}{2.938648in}}%
\pgfpathlineto{\pgfqpoint{0.827906in}{2.764542in}}%
\pgfpathlineto{\pgfqpoint{0.827906in}{2.764542in}}%
\pgfpathclose%
\pgfusepath{fill}%
\end{pgfscope}%
\begin{pgfscope}%
\pgfpathrectangle{\pgfqpoint{0.705073in}{0.712467in}}{\pgfqpoint{2.579497in}{3.165712in}}%
\pgfusepath{clip}%
\pgfsetbuttcap%
\pgfsetroundjoin%
\definecolor{currentfill}{rgb}{1.000000,0.498039,0.054902}%
\pgfsetfillcolor{currentfill}%
\pgfsetfillopacity{0.250000}%
\pgfsetlinewidth{0.000000pt}%
\definecolor{currentstroke}{rgb}{0.000000,0.000000,0.000000}%
\pgfsetstrokecolor{currentstroke}%
\pgfsetdash{}{0pt}%
\pgfpathmoveto{\pgfqpoint{0.827906in}{2.076436in}}%
\pgfpathlineto{\pgfqpoint{0.827906in}{1.674817in}}%
\pgfpathlineto{\pgfqpoint{1.646794in}{2.548100in}}%
\pgfpathlineto{\pgfqpoint{3.284570in}{3.336342in}}%
\pgfpathlineto{\pgfqpoint{3.284570in}{3.460709in}}%
\pgfpathlineto{\pgfqpoint{3.284570in}{3.460709in}}%
\pgfpathlineto{\pgfqpoint{1.646794in}{2.642112in}}%
\pgfpathlineto{\pgfqpoint{0.827906in}{2.076436in}}%
\pgfpathlineto{\pgfqpoint{0.827906in}{2.076436in}}%
\pgfpathclose%
\pgfusepath{fill}%
\end{pgfscope}%
\begin{pgfscope}%
\pgfpathrectangle{\pgfqpoint{0.705073in}{0.712467in}}{\pgfqpoint{2.579497in}{3.165712in}}%
\pgfusepath{clip}%
\pgfsetbuttcap%
\pgfsetroundjoin%
\definecolor{currentfill}{rgb}{0.172549,0.627451,0.172549}%
\pgfsetfillcolor{currentfill}%
\pgfsetfillopacity{0.250000}%
\pgfsetlinewidth{0.000000pt}%
\definecolor{currentstroke}{rgb}{0.000000,0.000000,0.000000}%
\pgfsetstrokecolor{currentstroke}%
\pgfsetdash{}{0pt}%
\pgfpathmoveto{\pgfqpoint{0.827906in}{1.549995in}}%
\pgfpathlineto{\pgfqpoint{0.827906in}{0.978142in}}%
\pgfpathlineto{\pgfqpoint{1.646794in}{2.082357in}}%
\pgfpathlineto{\pgfqpoint{3.284570in}{3.022341in}}%
\pgfpathlineto{\pgfqpoint{3.284570in}{3.271074in}}%
\pgfpathlineto{\pgfqpoint{3.284570in}{3.271074in}}%
\pgfpathlineto{\pgfqpoint{1.646794in}{2.364393in}}%
\pgfpathlineto{\pgfqpoint{0.827906in}{1.549995in}}%
\pgfpathlineto{\pgfqpoint{0.827906in}{1.549995in}}%
\pgfpathclose%
\pgfusepath{fill}%
\end{pgfscope}%
\begin{pgfscope}%
\pgfsetbuttcap%
\pgfsetroundjoin%
\definecolor{currentfill}{rgb}{0.000000,0.000000,0.000000}%
\pgfsetfillcolor{currentfill}%
\pgfsetlinewidth{0.803000pt}%
\definecolor{currentstroke}{rgb}{0.000000,0.000000,0.000000}%
\pgfsetstrokecolor{currentstroke}%
\pgfsetdash{}{0pt}%
\pgfsys@defobject{currentmarker}{\pgfqpoint{0.000000in}{-0.048611in}}{\pgfqpoint{0.000000in}{0.000000in}}{%
\pgfpathmoveto{\pgfqpoint{0.000000in}{0.000000in}}%
\pgfpathlineto{\pgfqpoint{0.000000in}{-0.048611in}}%
\pgfusepath{stroke,fill}%
}%
\begin{pgfscope}%
\pgfsys@transformshift{0.827906in}{0.712467in}%
\pgfsys@useobject{currentmarker}{}%
\end{pgfscope}%
\end{pgfscope}%
\begin{pgfscope}%
\definecolor{textcolor}{rgb}{0.000000,0.000000,0.000000}%
\pgfsetstrokecolor{textcolor}%
\pgfsetfillcolor{textcolor}%
\pgftext[x=0.827906in,y=0.615245in,,top]{\color{textcolor}\fontsize{14.000000}{16.800000}\selectfont \(\displaystyle {35}\)}%
\end{pgfscope}%
\begin{pgfscope}%
\pgfsetbuttcap%
\pgfsetroundjoin%
\definecolor{currentfill}{rgb}{0.000000,0.000000,0.000000}%
\pgfsetfillcolor{currentfill}%
\pgfsetlinewidth{0.803000pt}%
\definecolor{currentstroke}{rgb}{0.000000,0.000000,0.000000}%
\pgfsetstrokecolor{currentstroke}%
\pgfsetdash{}{0pt}%
\pgfsys@defobject{currentmarker}{\pgfqpoint{0.000000in}{-0.048611in}}{\pgfqpoint{0.000000in}{0.000000in}}{%
\pgfpathmoveto{\pgfqpoint{0.000000in}{0.000000in}}%
\pgfpathlineto{\pgfqpoint{0.000000in}{-0.048611in}}%
\pgfusepath{stroke,fill}%
}%
\begin{pgfscope}%
\pgfsys@transformshift{1.646794in}{0.712467in}%
\pgfsys@useobject{currentmarker}{}%
\end{pgfscope}%
\end{pgfscope}%
\begin{pgfscope}%
\definecolor{textcolor}{rgb}{0.000000,0.000000,0.000000}%
\pgfsetstrokecolor{textcolor}%
\pgfsetfillcolor{textcolor}%
\pgftext[x=1.646794in,y=0.615245in,,top]{\color{textcolor}\fontsize{14.000000}{16.800000}\selectfont \(\displaystyle {70}\)}%
\end{pgfscope}%
\begin{pgfscope}%
\pgfsetbuttcap%
\pgfsetroundjoin%
\definecolor{currentfill}{rgb}{0.000000,0.000000,0.000000}%
\pgfsetfillcolor{currentfill}%
\pgfsetlinewidth{0.803000pt}%
\definecolor{currentstroke}{rgb}{0.000000,0.000000,0.000000}%
\pgfsetstrokecolor{currentstroke}%
\pgfsetdash{}{0pt}%
\pgfsys@defobject{currentmarker}{\pgfqpoint{0.000000in}{-0.048611in}}{\pgfqpoint{0.000000in}{0.000000in}}{%
\pgfpathmoveto{\pgfqpoint{0.000000in}{0.000000in}}%
\pgfpathlineto{\pgfqpoint{0.000000in}{-0.048611in}}%
\pgfusepath{stroke,fill}%
}%
\begin{pgfscope}%
\pgfsys@transformshift{3.284570in}{0.712467in}%
\pgfsys@useobject{currentmarker}{}%
\end{pgfscope}%
\end{pgfscope}%
\begin{pgfscope}%
\definecolor{textcolor}{rgb}{0.000000,0.000000,0.000000}%
\pgfsetstrokecolor{textcolor}%
\pgfsetfillcolor{textcolor}%
\pgftext[x=3.284570in,y=0.615245in,,top]{\color{textcolor}\fontsize{14.000000}{16.800000}\selectfont \(\displaystyle {140}\)}%
\end{pgfscope}%
\begin{pgfscope}%
\definecolor{textcolor}{rgb}{0.000000,0.000000,0.000000}%
\pgfsetstrokecolor{textcolor}%
\pgfsetfillcolor{textcolor}%
\pgftext[x=1.994822in,y=0.288178in,,top]{\color{textcolor}\fontsize{14.000000}{16.800000}\selectfont demos}%
\end{pgfscope}%
\begin{pgfscope}%
\pgfsetbuttcap%
\pgfsetroundjoin%
\definecolor{currentfill}{rgb}{0.000000,0.000000,0.000000}%
\pgfsetfillcolor{currentfill}%
\pgfsetlinewidth{0.803000pt}%
\definecolor{currentstroke}{rgb}{0.000000,0.000000,0.000000}%
\pgfsetstrokecolor{currentstroke}%
\pgfsetdash{}{0pt}%
\pgfsys@defobject{currentmarker}{\pgfqpoint{-0.048611in}{0.000000in}}{\pgfqpoint{-0.000000in}{0.000000in}}{%
\pgfpathmoveto{\pgfqpoint{-0.000000in}{0.000000in}}%
\pgfpathlineto{\pgfqpoint{-0.048611in}{0.000000in}}%
\pgfusepath{stroke,fill}%
}%
\begin{pgfscope}%
\pgfsys@transformshift{0.705073in}{0.856363in}%
\pgfsys@useobject{currentmarker}{}%
\end{pgfscope}%
\end{pgfscope}%
\begin{pgfscope}%
\definecolor{textcolor}{rgb}{0.000000,0.000000,0.000000}%
\pgfsetstrokecolor{textcolor}%
\pgfsetfillcolor{textcolor}%
\pgftext[x=0.357623in, y=0.782497in, left, base]{\color{textcolor}\fontsize{14.000000}{16.800000}\selectfont \(\displaystyle {0.4}\)}%
\end{pgfscope}%
\begin{pgfscope}%
\pgfsetbuttcap%
\pgfsetroundjoin%
\definecolor{currentfill}{rgb}{0.000000,0.000000,0.000000}%
\pgfsetfillcolor{currentfill}%
\pgfsetlinewidth{0.803000pt}%
\definecolor{currentstroke}{rgb}{0.000000,0.000000,0.000000}%
\pgfsetstrokecolor{currentstroke}%
\pgfsetdash{}{0pt}%
\pgfsys@defobject{currentmarker}{\pgfqpoint{-0.048611in}{0.000000in}}{\pgfqpoint{-0.000000in}{0.000000in}}{%
\pgfpathmoveto{\pgfqpoint{-0.000000in}{0.000000in}}%
\pgfpathlineto{\pgfqpoint{-0.048611in}{0.000000in}}%
\pgfusepath{stroke,fill}%
}%
\begin{pgfscope}%
\pgfsys@transformshift{0.705073in}{2.295323in}%
\pgfsys@useobject{currentmarker}{}%
\end{pgfscope}%
\end{pgfscope}%
\begin{pgfscope}%
\definecolor{textcolor}{rgb}{0.000000,0.000000,0.000000}%
\pgfsetstrokecolor{textcolor}%
\pgfsetfillcolor{textcolor}%
\pgftext[x=0.357623in, y=2.221457in, left, base]{\color{textcolor}\fontsize{14.000000}{16.800000}\selectfont \(\displaystyle {0.6}\)}%
\end{pgfscope}%
\begin{pgfscope}%
\pgfsetbuttcap%
\pgfsetroundjoin%
\definecolor{currentfill}{rgb}{0.000000,0.000000,0.000000}%
\pgfsetfillcolor{currentfill}%
\pgfsetlinewidth{0.803000pt}%
\definecolor{currentstroke}{rgb}{0.000000,0.000000,0.000000}%
\pgfsetstrokecolor{currentstroke}%
\pgfsetdash{}{0pt}%
\pgfsys@defobject{currentmarker}{\pgfqpoint{-0.048611in}{0.000000in}}{\pgfqpoint{-0.000000in}{0.000000in}}{%
\pgfpathmoveto{\pgfqpoint{-0.000000in}{0.000000in}}%
\pgfpathlineto{\pgfqpoint{-0.048611in}{0.000000in}}%
\pgfusepath{stroke,fill}%
}%
\begin{pgfscope}%
\pgfsys@transformshift{0.705073in}{3.734283in}%
\pgfsys@useobject{currentmarker}{}%
\end{pgfscope}%
\end{pgfscope}%
\begin{pgfscope}%
\definecolor{textcolor}{rgb}{0.000000,0.000000,0.000000}%
\pgfsetstrokecolor{textcolor}%
\pgfsetfillcolor{textcolor}%
\pgftext[x=0.357623in, y=3.660417in, left, base]{\color{textcolor}\fontsize{14.000000}{16.800000}\selectfont \(\displaystyle {0.8}\)}%
\end{pgfscope}%
\begin{pgfscope}%
\definecolor{textcolor}{rgb}{0.000000,0.000000,0.000000}%
\pgfsetstrokecolor{textcolor}%
\pgfsetfillcolor{textcolor}%
\pgftext[x=0.288178in,y=2.295323in,,bottom,rotate=90.000000]{\color{textcolor}\fontsize{14.000000}{16.800000}\selectfont success rate}%
\end{pgfscope}%
\begin{pgfscope}%
\pgfpathrectangle{\pgfqpoint{0.705073in}{0.712467in}}{\pgfqpoint{2.579497in}{3.165712in}}%
\pgfusepath{clip}%
\pgfsetrectcap%
\pgfsetroundjoin%
\pgfsetlinewidth{1.505625pt}%
\definecolor{currentstroke}{rgb}{0.121569,0.466667,0.705882}%
\pgfsetstrokecolor{currentstroke}%
\pgfsetdash{}{0pt}%
\pgfpathmoveto{\pgfqpoint{0.827906in}{2.187401in}}%
\pgfpathlineto{\pgfqpoint{1.646794in}{2.607098in}}%
\pgfpathlineto{\pgfqpoint{3.284570in}{3.578395in}}%
\pgfusepath{stroke}%
\end{pgfscope}%
\begin{pgfscope}%
\pgfpathrectangle{\pgfqpoint{0.705073in}{0.712467in}}{\pgfqpoint{2.579497in}{3.165712in}}%
\pgfusepath{clip}%
\pgfsetrectcap%
\pgfsetroundjoin%
\pgfsetlinewidth{1.505625pt}%
\definecolor{currentstroke}{rgb}{1.000000,0.498039,0.054902}%
\pgfsetstrokecolor{currentstroke}%
\pgfsetdash{}{0pt}%
\pgfpathmoveto{\pgfqpoint{0.827906in}{1.875626in}}%
\pgfpathlineto{\pgfqpoint{1.646794in}{2.595106in}}%
\pgfpathlineto{\pgfqpoint{3.284570in}{3.398525in}}%
\pgfusepath{stroke}%
\end{pgfscope}%
\begin{pgfscope}%
\pgfpathrectangle{\pgfqpoint{0.705073in}{0.712467in}}{\pgfqpoint{2.579497in}{3.165712in}}%
\pgfusepath{clip}%
\pgfsetrectcap%
\pgfsetroundjoin%
\pgfsetlinewidth{1.505625pt}%
\definecolor{currentstroke}{rgb}{0.172549,0.627451,0.172549}%
\pgfsetstrokecolor{currentstroke}%
\pgfsetdash{}{0pt}%
\pgfpathmoveto{\pgfqpoint{0.827906in}{1.264068in}}%
\pgfpathlineto{\pgfqpoint{1.646794in}{2.223375in}}%
\pgfpathlineto{\pgfqpoint{3.284570in}{3.146708in}}%
\pgfusepath{stroke}%
\end{pgfscope}%
\begin{pgfscope}%
\pgfsetrectcap%
\pgfsetmiterjoin%
\pgfsetlinewidth{0.803000pt}%
\definecolor{currentstroke}{rgb}{0.000000,0.000000,0.000000}%
\pgfsetstrokecolor{currentstroke}%
\pgfsetdash{}{0pt}%
\pgfpathmoveto{\pgfqpoint{0.705073in}{0.856363in}}%
\pgfpathlineto{\pgfqpoint{0.705073in}{3.734283in}}%
\pgfusepath{stroke}%
\end{pgfscope}%
\begin{pgfscope}%
\pgfsetrectcap%
\pgfsetmiterjoin%
\pgfsetlinewidth{0.803000pt}%
\definecolor{currentstroke}{rgb}{0.000000,0.000000,0.000000}%
\pgfsetstrokecolor{currentstroke}%
\pgfsetdash{}{0pt}%
\pgfpathmoveto{\pgfqpoint{0.827906in}{0.712467in}}%
\pgfpathlineto{\pgfqpoint{3.284570in}{0.712467in}}%
\pgfusepath{stroke}%
\end{pgfscope}%
\begin{pgfscope}%
\definecolor{textcolor}{rgb}{0.000000,0.000000,0.000000}%
\pgfsetstrokecolor{textcolor}%
\pgfsetfillcolor{textcolor}%
\pgftext[x=1.994822in,y=3.961512in,,base]{\color{textcolor}\fontsize{16.800000}{20.160000}\selectfont Phone}%
\end{pgfscope}%
\begin{pgfscope}%
\pgfsetbuttcap%
\pgfsetmiterjoin%
\pgfsetlinewidth{0.000000pt}%
\definecolor{currentstroke}{rgb}{0.000000,0.000000,0.000000}%
\pgfsetstrokecolor{currentstroke}%
\pgfsetstrokeopacity{0.000000}%
\pgfsetdash{}{0pt}%
\pgfpathmoveto{\pgfqpoint{3.861266in}{0.712467in}}%
\pgfpathlineto{\pgfqpoint{6.440763in}{0.712467in}}%
\pgfpathlineto{\pgfqpoint{6.440763in}{3.878179in}}%
\pgfpathlineto{\pgfqpoint{3.861266in}{3.878179in}}%
\pgfpathlineto{\pgfqpoint{3.861266in}{0.712467in}}%
\pgfpathclose%
\pgfusepath{}%
\end{pgfscope}%
\begin{pgfscope}%
\pgfpathrectangle{\pgfqpoint{3.861266in}{0.712467in}}{\pgfqpoint{2.579497in}{3.165712in}}%
\pgfusepath{clip}%
\pgfsetbuttcap%
\pgfsetroundjoin%
\definecolor{currentfill}{rgb}{0.121569,0.466667,0.705882}%
\pgfsetfillcolor{currentfill}%
\pgfsetfillopacity{0.250000}%
\pgfsetlinewidth{0.000000pt}%
\definecolor{currentstroke}{rgb}{0.000000,0.000000,0.000000}%
\pgfsetstrokecolor{currentstroke}%
\pgfsetdash{}{0pt}%
\pgfpathmoveto{\pgfqpoint{3.984099in}{1.932294in}}%
\pgfpathlineto{\pgfqpoint{3.984099in}{1.200206in}}%
\pgfpathlineto{\pgfqpoint{4.802987in}{2.782265in}}%
\pgfpathlineto{\pgfqpoint{6.440763in}{3.157416in}}%
\pgfpathlineto{\pgfqpoint{6.440763in}{3.428588in}}%
\pgfpathlineto{\pgfqpoint{6.440763in}{3.428588in}}%
\pgfpathlineto{\pgfqpoint{4.802987in}{3.017107in}}%
\pgfpathlineto{\pgfqpoint{3.984099in}{1.932294in}}%
\pgfpathlineto{\pgfqpoint{3.984099in}{1.932294in}}%
\pgfpathclose%
\pgfusepath{fill}%
\end{pgfscope}%
\begin{pgfscope}%
\pgfpathrectangle{\pgfqpoint{3.861266in}{0.712467in}}{\pgfqpoint{2.579497in}{3.165712in}}%
\pgfusepath{clip}%
\pgfsetbuttcap%
\pgfsetroundjoin%
\definecolor{currentfill}{rgb}{1.000000,0.498039,0.054902}%
\pgfsetfillcolor{currentfill}%
\pgfsetfillopacity{0.250000}%
\pgfsetlinewidth{0.000000pt}%
\definecolor{currentstroke}{rgb}{0.000000,0.000000,0.000000}%
\pgfsetstrokecolor{currentstroke}%
\pgfsetdash{}{0pt}%
\pgfpathmoveto{\pgfqpoint{3.984099in}{2.736228in}}%
\pgfpathlineto{\pgfqpoint{3.984099in}{1.796860in}}%
\pgfpathlineto{\pgfqpoint{4.802987in}{2.809786in}}%
\pgfpathlineto{\pgfqpoint{6.440763in}{2.662688in}}%
\pgfpathlineto{\pgfqpoint{6.440763in}{3.155870in}}%
\pgfpathlineto{\pgfqpoint{6.440763in}{3.155870in}}%
\pgfpathlineto{\pgfqpoint{4.802987in}{3.008772in}}%
\pgfpathlineto{\pgfqpoint{3.984099in}{2.736228in}}%
\pgfpathlineto{\pgfqpoint{3.984099in}{2.736228in}}%
\pgfpathclose%
\pgfusepath{fill}%
\end{pgfscope}%
\begin{pgfscope}%
\pgfpathrectangle{\pgfqpoint{3.861266in}{0.712467in}}{\pgfqpoint{2.579497in}{3.165712in}}%
\pgfusepath{clip}%
\pgfsetbuttcap%
\pgfsetroundjoin%
\definecolor{currentfill}{rgb}{0.172549,0.627451,0.172549}%
\pgfsetfillcolor{currentfill}%
\pgfsetfillopacity{0.250000}%
\pgfsetlinewidth{0.000000pt}%
\definecolor{currentstroke}{rgb}{0.000000,0.000000,0.000000}%
\pgfsetstrokecolor{currentstroke}%
\pgfsetdash{}{0pt}%
\pgfpathmoveto{\pgfqpoint{3.984099in}{1.075589in}}%
\pgfpathlineto{\pgfqpoint{3.984099in}{0.828998in}}%
\pgfpathlineto{\pgfqpoint{4.802987in}{1.478143in}}%
\pgfpathlineto{\pgfqpoint{6.440763in}{1.840200in}}%
\pgfpathlineto{\pgfqpoint{6.440763in}{2.520213in}}%
\pgfpathlineto{\pgfqpoint{6.440763in}{2.520213in}}%
\pgfpathlineto{\pgfqpoint{4.802987in}{1.673543in}}%
\pgfpathlineto{\pgfqpoint{3.984099in}{1.075589in}}%
\pgfpathlineto{\pgfqpoint{3.984099in}{1.075589in}}%
\pgfpathclose%
\pgfusepath{fill}%
\end{pgfscope}%
\begin{pgfscope}%
\pgfsetbuttcap%
\pgfsetroundjoin%
\definecolor{currentfill}{rgb}{0.000000,0.000000,0.000000}%
\pgfsetfillcolor{currentfill}%
\pgfsetlinewidth{0.803000pt}%
\definecolor{currentstroke}{rgb}{0.000000,0.000000,0.000000}%
\pgfsetstrokecolor{currentstroke}%
\pgfsetdash{}{0pt}%
\pgfsys@defobject{currentmarker}{\pgfqpoint{0.000000in}{-0.048611in}}{\pgfqpoint{0.000000in}{0.000000in}}{%
\pgfpathmoveto{\pgfqpoint{0.000000in}{0.000000in}}%
\pgfpathlineto{\pgfqpoint{0.000000in}{-0.048611in}}%
\pgfusepath{stroke,fill}%
}%
\begin{pgfscope}%
\pgfsys@transformshift{3.984099in}{0.712467in}%
\pgfsys@useobject{currentmarker}{}%
\end{pgfscope}%
\end{pgfscope}%
\begin{pgfscope}%
\definecolor{textcolor}{rgb}{0.000000,0.000000,0.000000}%
\pgfsetstrokecolor{textcolor}%
\pgfsetfillcolor{textcolor}%
\pgftext[x=3.984099in,y=0.615245in,,top]{\color{textcolor}\fontsize{14.000000}{16.800000}\selectfont \(\displaystyle {35}\)}%
\end{pgfscope}%
\begin{pgfscope}%
\pgfsetbuttcap%
\pgfsetroundjoin%
\definecolor{currentfill}{rgb}{0.000000,0.000000,0.000000}%
\pgfsetfillcolor{currentfill}%
\pgfsetlinewidth{0.803000pt}%
\definecolor{currentstroke}{rgb}{0.000000,0.000000,0.000000}%
\pgfsetstrokecolor{currentstroke}%
\pgfsetdash{}{0pt}%
\pgfsys@defobject{currentmarker}{\pgfqpoint{0.000000in}{-0.048611in}}{\pgfqpoint{0.000000in}{0.000000in}}{%
\pgfpathmoveto{\pgfqpoint{0.000000in}{0.000000in}}%
\pgfpathlineto{\pgfqpoint{0.000000in}{-0.048611in}}%
\pgfusepath{stroke,fill}%
}%
\begin{pgfscope}%
\pgfsys@transformshift{4.802987in}{0.712467in}%
\pgfsys@useobject{currentmarker}{}%
\end{pgfscope}%
\end{pgfscope}%
\begin{pgfscope}%
\definecolor{textcolor}{rgb}{0.000000,0.000000,0.000000}%
\pgfsetstrokecolor{textcolor}%
\pgfsetfillcolor{textcolor}%
\pgftext[x=4.802987in,y=0.615245in,,top]{\color{textcolor}\fontsize{14.000000}{16.800000}\selectfont \(\displaystyle {70}\)}%
\end{pgfscope}%
\begin{pgfscope}%
\pgfsetbuttcap%
\pgfsetroundjoin%
\definecolor{currentfill}{rgb}{0.000000,0.000000,0.000000}%
\pgfsetfillcolor{currentfill}%
\pgfsetlinewidth{0.803000pt}%
\definecolor{currentstroke}{rgb}{0.000000,0.000000,0.000000}%
\pgfsetstrokecolor{currentstroke}%
\pgfsetdash{}{0pt}%
\pgfsys@defobject{currentmarker}{\pgfqpoint{0.000000in}{-0.048611in}}{\pgfqpoint{0.000000in}{0.000000in}}{%
\pgfpathmoveto{\pgfqpoint{0.000000in}{0.000000in}}%
\pgfpathlineto{\pgfqpoint{0.000000in}{-0.048611in}}%
\pgfusepath{stroke,fill}%
}%
\begin{pgfscope}%
\pgfsys@transformshift{6.440763in}{0.712467in}%
\pgfsys@useobject{currentmarker}{}%
\end{pgfscope}%
\end{pgfscope}%
\begin{pgfscope}%
\definecolor{textcolor}{rgb}{0.000000,0.000000,0.000000}%
\pgfsetstrokecolor{textcolor}%
\pgfsetfillcolor{textcolor}%
\pgftext[x=6.440763in,y=0.615245in,,top]{\color{textcolor}\fontsize{14.000000}{16.800000}\selectfont \(\displaystyle {140}\)}%
\end{pgfscope}%
\begin{pgfscope}%
\definecolor{textcolor}{rgb}{0.000000,0.000000,0.000000}%
\pgfsetstrokecolor{textcolor}%
\pgfsetfillcolor{textcolor}%
\pgftext[x=5.151014in,y=0.288178in,,top]{\color{textcolor}\fontsize{14.000000}{16.800000}\selectfont demos}%
\end{pgfscope}%
\begin{pgfscope}%
\pgfsetbuttcap%
\pgfsetroundjoin%
\definecolor{currentfill}{rgb}{0.000000,0.000000,0.000000}%
\pgfsetfillcolor{currentfill}%
\pgfsetlinewidth{0.803000pt}%
\definecolor{currentstroke}{rgb}{0.000000,0.000000,0.000000}%
\pgfsetstrokecolor{currentstroke}%
\pgfsetdash{}{0pt}%
\pgfsys@defobject{currentmarker}{\pgfqpoint{-0.048611in}{0.000000in}}{\pgfqpoint{-0.000000in}{0.000000in}}{%
\pgfpathmoveto{\pgfqpoint{-0.000000in}{0.000000in}}%
\pgfpathlineto{\pgfqpoint{-0.048611in}{0.000000in}}%
\pgfusepath{stroke,fill}%
}%
\begin{pgfscope}%
\pgfsys@transformshift{3.861266in}{0.856363in}%
\pgfsys@useobject{currentmarker}{}%
\end{pgfscope}%
\end{pgfscope}%
\begin{pgfscope}%
\definecolor{textcolor}{rgb}{0.000000,0.000000,0.000000}%
\pgfsetstrokecolor{textcolor}%
\pgfsetfillcolor{textcolor}%
\pgftext[x=3.415900in, y=0.782497in, left, base]{\color{textcolor}\fontsize{14.000000}{16.800000}\selectfont \(\displaystyle {0.50}\)}%
\end{pgfscope}%
\begin{pgfscope}%
\pgfsetbuttcap%
\pgfsetroundjoin%
\definecolor{currentfill}{rgb}{0.000000,0.000000,0.000000}%
\pgfsetfillcolor{currentfill}%
\pgfsetlinewidth{0.803000pt}%
\definecolor{currentstroke}{rgb}{0.000000,0.000000,0.000000}%
\pgfsetstrokecolor{currentstroke}%
\pgfsetdash{}{0pt}%
\pgfsys@defobject{currentmarker}{\pgfqpoint{-0.048611in}{0.000000in}}{\pgfqpoint{-0.000000in}{0.000000in}}{%
\pgfpathmoveto{\pgfqpoint{-0.000000in}{0.000000in}}%
\pgfpathlineto{\pgfqpoint{-0.048611in}{0.000000in}}%
\pgfusepath{stroke,fill}%
}%
\begin{pgfscope}%
\pgfsys@transformshift{3.861266in}{2.295323in}%
\pgfsys@useobject{currentmarker}{}%
\end{pgfscope}%
\end{pgfscope}%
\begin{pgfscope}%
\definecolor{textcolor}{rgb}{0.000000,0.000000,0.000000}%
\pgfsetstrokecolor{textcolor}%
\pgfsetfillcolor{textcolor}%
\pgftext[x=3.415900in, y=2.221457in, left, base]{\color{textcolor}\fontsize{14.000000}{16.800000}\selectfont \(\displaystyle {0.75}\)}%
\end{pgfscope}%
\begin{pgfscope}%
\pgfsetbuttcap%
\pgfsetroundjoin%
\definecolor{currentfill}{rgb}{0.000000,0.000000,0.000000}%
\pgfsetfillcolor{currentfill}%
\pgfsetlinewidth{0.803000pt}%
\definecolor{currentstroke}{rgb}{0.000000,0.000000,0.000000}%
\pgfsetstrokecolor{currentstroke}%
\pgfsetdash{}{0pt}%
\pgfsys@defobject{currentmarker}{\pgfqpoint{-0.048611in}{0.000000in}}{\pgfqpoint{-0.000000in}{0.000000in}}{%
\pgfpathmoveto{\pgfqpoint{-0.000000in}{0.000000in}}%
\pgfpathlineto{\pgfqpoint{-0.048611in}{0.000000in}}%
\pgfusepath{stroke,fill}%
}%
\begin{pgfscope}%
\pgfsys@transformshift{3.861266in}{3.734283in}%
\pgfsys@useobject{currentmarker}{}%
\end{pgfscope}%
\end{pgfscope}%
\begin{pgfscope}%
\definecolor{textcolor}{rgb}{0.000000,0.000000,0.000000}%
\pgfsetstrokecolor{textcolor}%
\pgfsetfillcolor{textcolor}%
\pgftext[x=3.415900in, y=3.660417in, left, base]{\color{textcolor}\fontsize{14.000000}{16.800000}\selectfont \(\displaystyle {1.00}\)}%
\end{pgfscope}%
\begin{pgfscope}%
\pgfpathrectangle{\pgfqpoint{3.861266in}{0.712467in}}{\pgfqpoint{2.579497in}{3.165712in}}%
\pgfusepath{clip}%
\pgfsetrectcap%
\pgfsetroundjoin%
\pgfsetlinewidth{1.505625pt}%
\definecolor{currentstroke}{rgb}{0.121569,0.466667,0.705882}%
\pgfsetstrokecolor{currentstroke}%
\pgfsetdash{}{0pt}%
\pgfpathmoveto{\pgfqpoint{3.984099in}{1.566250in}}%
\pgfpathlineto{\pgfqpoint{4.802987in}{2.899686in}}%
\pgfpathlineto{\pgfqpoint{6.440763in}{3.293002in}}%
\pgfusepath{stroke}%
\end{pgfscope}%
\begin{pgfscope}%
\pgfpathrectangle{\pgfqpoint{3.861266in}{0.712467in}}{\pgfqpoint{2.579497in}{3.165712in}}%
\pgfusepath{clip}%
\pgfsetrectcap%
\pgfsetroundjoin%
\pgfsetlinewidth{1.505625pt}%
\definecolor{currentstroke}{rgb}{1.000000,0.498039,0.054902}%
\pgfsetstrokecolor{currentstroke}%
\pgfsetdash{}{0pt}%
\pgfpathmoveto{\pgfqpoint{3.984099in}{2.266544in}}%
\pgfpathlineto{\pgfqpoint{4.802987in}{2.909279in}}%
\pgfpathlineto{\pgfqpoint{6.440763in}{2.909279in}}%
\pgfusepath{stroke}%
\end{pgfscope}%
\begin{pgfscope}%
\pgfpathrectangle{\pgfqpoint{3.861266in}{0.712467in}}{\pgfqpoint{2.579497in}{3.165712in}}%
\pgfusepath{clip}%
\pgfsetrectcap%
\pgfsetroundjoin%
\pgfsetlinewidth{1.505625pt}%
\definecolor{currentstroke}{rgb}{0.172549,0.627451,0.172549}%
\pgfsetstrokecolor{currentstroke}%
\pgfsetdash{}{0pt}%
\pgfpathmoveto{\pgfqpoint{3.984099in}{0.952294in}}%
\pgfpathlineto{\pgfqpoint{4.802987in}{1.575843in}}%
\pgfpathlineto{\pgfqpoint{6.440763in}{2.180206in}}%
\pgfusepath{stroke}%
\end{pgfscope}%
\begin{pgfscope}%
\pgfsetrectcap%
\pgfsetmiterjoin%
\pgfsetlinewidth{0.803000pt}%
\definecolor{currentstroke}{rgb}{0.000000,0.000000,0.000000}%
\pgfsetstrokecolor{currentstroke}%
\pgfsetdash{}{0pt}%
\pgfpathmoveto{\pgfqpoint{3.861266in}{0.856363in}}%
\pgfpathlineto{\pgfqpoint{3.861266in}{3.734283in}}%
\pgfusepath{stroke}%
\end{pgfscope}%
\begin{pgfscope}%
\pgfsetrectcap%
\pgfsetmiterjoin%
\pgfsetlinewidth{0.803000pt}%
\definecolor{currentstroke}{rgb}{0.000000,0.000000,0.000000}%
\pgfsetstrokecolor{currentstroke}%
\pgfsetdash{}{0pt}%
\pgfpathmoveto{\pgfqpoint{3.984099in}{0.712467in}}%
\pgfpathlineto{\pgfqpoint{6.440763in}{0.712467in}}%
\pgfusepath{stroke}%
\end{pgfscope}%
\begin{pgfscope}%
\definecolor{textcolor}{rgb}{0.000000,0.000000,0.000000}%
\pgfsetstrokecolor{textcolor}%
\pgfsetfillcolor{textcolor}%
\pgftext[x=5.151014in,y=3.961512in,,base]{\color{textcolor}\fontsize{16.800000}{20.160000}\selectfont Rubbish}%
\end{pgfscope}%
\begin{pgfscope}%
\pgfsetrectcap%
\pgfsetroundjoin%
\pgfsetlinewidth{4.015000pt}%
\definecolor{currentstroke}{rgb}{0.121569,0.466667,0.705882}%
\pgfsetstrokecolor{currentstroke}%
\pgfsetdash{}{0pt}%
\pgfpathmoveto{\pgfqpoint{0.298024in}{4.464157in}}%
\pgfpathlineto{\pgfqpoint{0.492469in}{4.464157in}}%
\pgfpathlineto{\pgfqpoint{0.686913in}{4.464157in}}%
\pgfusepath{stroke}%
\end{pgfscope}%
\begin{pgfscope}%
\definecolor{textcolor}{rgb}{0.000000,0.000000,0.000000}%
\pgfsetstrokecolor{textcolor}%
\pgfsetfillcolor{textcolor}%
\pgftext[x=0.842469in,y=4.396102in,left,base]{\color{textcolor}\fontsize{14.000000}{16.800000}\selectfont Keypoints}%
\end{pgfscope}%
\begin{pgfscope}%
\pgfsetrectcap%
\pgfsetroundjoin%
\pgfsetlinewidth{4.015000pt}%
\definecolor{currentstroke}{rgb}{1.000000,0.498039,0.054902}%
\pgfsetstrokecolor{currentstroke}%
\pgfsetdash{}{0pt}%
\pgfpathmoveto{\pgfqpoint{2.203200in}{4.464157in}}%
\pgfpathlineto{\pgfqpoint{2.397645in}{4.464157in}}%
\pgfpathlineto{\pgfqpoint{2.592089in}{4.464157in}}%
\pgfusepath{stroke}%
\end{pgfscope}%
\begin{pgfscope}%
\definecolor{textcolor}{rgb}{0.000000,0.000000,0.000000}%
\pgfsetstrokecolor{textcolor}%
\pgfsetfillcolor{textcolor}%
\pgftext[x=2.747645in,y=4.396102in,left,base]{\color{textcolor}\fontsize{14.000000}{16.800000}\selectfont Filter}%
\end{pgfscope}%
\begin{pgfscope}%
\pgfsetrectcap%
\pgfsetroundjoin%
\pgfsetlinewidth{4.015000pt}%
\definecolor{currentstroke}{rgb}{0.172549,0.627451,0.172549}%
\pgfsetstrokecolor{currentstroke}%
\pgfsetdash{}{0pt}%
\pgfpathmoveto{\pgfqpoint{3.681984in}{4.464157in}}%
\pgfpathlineto{\pgfqpoint{3.876429in}{4.464157in}}%
\pgfpathlineto{\pgfqpoint{4.070873in}{4.464157in}}%
\pgfusepath{stroke}%
\end{pgfscope}%
\begin{pgfscope}%
\definecolor{textcolor}{rgb}{0.000000,0.000000,0.000000}%
\pgfsetstrokecolor{textcolor}%
\pgfsetfillcolor{textcolor}%
\pgftext[x=4.226429in,y=4.396102in,left,base]{\color{textcolor}\fontsize{14.000000}{16.800000}\selectfont MONet}%
\end{pgfscope}%
\end{pgfpicture}%
\makeatother%
\endgroup%

%% file: figures/joint_dropout.pgf
\begingroup%
\makeatletter%
\begin{pgfpicture}%
\pgfpathrectangle{\pgfpointorigin}{\pgfqpoint{6.686953in}{4.238791in}}%
\pgfusepath{use as bounding box, clip}%
\begin{pgfscope}%
\pgfsetbuttcap%
\pgfsetmiterjoin%
\pgfsetlinewidth{0.000000pt}%
\definecolor{currentstroke}{rgb}{0.000000,0.000000,0.000000}%
\pgfsetstrokecolor{currentstroke}%
\pgfsetstrokeopacity{0.000000}%
\pgfsetdash{}{0pt}%
\pgfpathmoveto{\pgfqpoint{0.000000in}{0.000000in}}%
\pgfpathlineto{\pgfqpoint{6.686953in}{0.000000in}}%
\pgfpathlineto{\pgfqpoint{6.686953in}{4.238791in}}%
\pgfpathlineto{\pgfqpoint{0.000000in}{4.238791in}}%
\pgfpathlineto{\pgfqpoint{0.000000in}{0.000000in}}%
\pgfpathclose%
\pgfusepath{}%
\end{pgfscope}%
\begin{pgfscope}%
\pgfsetbuttcap%
\pgfsetmiterjoin%
\pgfsetlinewidth{0.000000pt}%
\definecolor{currentstroke}{rgb}{0.000000,0.000000,0.000000}%
\pgfsetstrokecolor{currentstroke}%
\pgfsetstrokeopacity{0.000000}%
\pgfsetdash{}{0pt}%
\pgfpathmoveto{\pgfqpoint{0.705073in}{0.712467in}}%
\pgfpathlineto{\pgfqpoint{3.353067in}{0.712467in}}%
\pgfpathlineto{\pgfqpoint{3.353067in}{3.878179in}}%
\pgfpathlineto{\pgfqpoint{0.705073in}{3.878179in}}%
\pgfpathlineto{\pgfqpoint{0.705073in}{0.712467in}}%
\pgfpathclose%
\pgfusepath{}%
\end{pgfscope}%
\begin{pgfscope}%
\pgfpathrectangle{\pgfqpoint{0.705073in}{0.712467in}}{\pgfqpoint{2.647994in}{3.165712in}}%
\pgfusepath{clip}%
\pgfsetbuttcap%
\pgfsetroundjoin%
\definecolor{currentfill}{rgb}{0.121569,0.466667,0.705882}%
\pgfsetfillcolor{currentfill}%
\pgfsetfillopacity{0.250000}%
\pgfsetlinewidth{0.000000pt}%
\definecolor{currentstroke}{rgb}{0.000000,0.000000,0.000000}%
\pgfsetstrokecolor{currentstroke}%
\pgfsetdash{}{0pt}%
\pgfpathmoveto{\pgfqpoint{0.831168in}{3.811058in}}%
\pgfpathlineto{\pgfqpoint{0.831168in}{3.449658in}}%
\pgfpathlineto{\pgfqpoint{2.092117in}{1.998846in}}%
\pgfpathlineto{\pgfqpoint{3.353067in}{0.553357in}}%
\pgfpathlineto{\pgfqpoint{3.353067in}{1.575069in}}%
\pgfpathlineto{\pgfqpoint{3.353067in}{1.575069in}}%
\pgfpathlineto{\pgfqpoint{2.092117in}{3.007500in}}%
\pgfpathlineto{\pgfqpoint{0.831168in}{3.811058in}}%
\pgfpathlineto{\pgfqpoint{0.831168in}{3.811058in}}%
\pgfpathclose%
\pgfusepath{fill}%
\end{pgfscope}%
\begin{pgfscope}%
\pgfpathrectangle{\pgfqpoint{0.705073in}{0.712467in}}{\pgfqpoint{2.647994in}{3.165712in}}%
\pgfusepath{clip}%
\pgfsetbuttcap%
\pgfsetroundjoin%
\definecolor{currentfill}{rgb}{1.000000,0.498039,0.054902}%
\pgfsetfillcolor{currentfill}%
\pgfsetfillopacity{0.250000}%
\pgfsetlinewidth{0.000000pt}%
\definecolor{currentstroke}{rgb}{0.000000,0.000000,0.000000}%
\pgfsetstrokecolor{currentstroke}%
\pgfsetdash{}{0pt}%
\pgfpathmoveto{\pgfqpoint{0.831168in}{3.551900in}}%
\pgfpathlineto{\pgfqpoint{0.831168in}{3.468989in}}%
\pgfpathlineto{\pgfqpoint{2.092117in}{2.131222in}}%
\pgfpathlineto{\pgfqpoint{3.353067in}{2.320905in}}%
\pgfpathlineto{\pgfqpoint{3.353067in}{2.477591in}}%
\pgfpathlineto{\pgfqpoint{3.353067in}{2.477591in}}%
\pgfpathlineto{\pgfqpoint{2.092117in}{2.891112in}}%
\pgfpathlineto{\pgfqpoint{0.831168in}{3.551900in}}%
\pgfpathlineto{\pgfqpoint{0.831168in}{3.551900in}}%
\pgfpathclose%
\pgfusepath{fill}%
\end{pgfscope}%
\begin{pgfscope}%
\pgfpathrectangle{\pgfqpoint{0.705073in}{0.712467in}}{\pgfqpoint{2.647994in}{3.165712in}}%
\pgfusepath{clip}%
\pgfsetbuttcap%
\pgfsetroundjoin%
\definecolor{currentfill}{rgb}{0.172549,0.627451,0.172549}%
\pgfsetfillcolor{currentfill}%
\pgfsetfillopacity{0.250000}%
\pgfsetlinewidth{0.000000pt}%
\definecolor{currentstroke}{rgb}{0.000000,0.000000,0.000000}%
\pgfsetstrokecolor{currentstroke}%
\pgfsetdash{}{0pt}%
\pgfpathmoveto{\pgfqpoint{0.831168in}{3.425477in}}%
\pgfpathlineto{\pgfqpoint{0.831168in}{3.259655in}}%
\pgfpathlineto{\pgfqpoint{2.092117in}{2.773372in}}%
\pgfpathlineto{\pgfqpoint{3.353067in}{2.391216in}}%
\pgfpathlineto{\pgfqpoint{3.353067in}{2.934898in}}%
\pgfpathlineto{\pgfqpoint{3.353067in}{2.934898in}}%
\pgfpathlineto{\pgfqpoint{2.092117in}{3.112338in}}%
\pgfpathlineto{\pgfqpoint{0.831168in}{3.425477in}}%
\pgfpathlineto{\pgfqpoint{0.831168in}{3.425477in}}%
\pgfpathclose%
\pgfusepath{fill}%
\end{pgfscope}%
\begin{pgfscope}%
\pgfsetbuttcap%
\pgfsetroundjoin%
\definecolor{currentfill}{rgb}{0.000000,0.000000,0.000000}%
\pgfsetfillcolor{currentfill}%
\pgfsetlinewidth{0.803000pt}%
\definecolor{currentstroke}{rgb}{0.000000,0.000000,0.000000}%
\pgfsetstrokecolor{currentstroke}%
\pgfsetdash{}{0pt}%
\pgfsys@defobject{currentmarker}{\pgfqpoint{0.000000in}{-0.048611in}}{\pgfqpoint{0.000000in}{0.000000in}}{%
\pgfpathmoveto{\pgfqpoint{0.000000in}{0.000000in}}%
\pgfpathlineto{\pgfqpoint{0.000000in}{-0.048611in}}%
\pgfusepath{stroke,fill}%
}%
\begin{pgfscope}%
\pgfsys@transformshift{0.831168in}{0.712467in}%
\pgfsys@useobject{currentmarker}{}%
\end{pgfscope}%
\end{pgfscope}%
\begin{pgfscope}%
\definecolor{textcolor}{rgb}{0.000000,0.000000,0.000000}%
\pgfsetstrokecolor{textcolor}%
\pgfsetfillcolor{textcolor}%
\pgftext[x=0.831168in,y=0.615245in,,top]{\color{textcolor}\fontsize{14.000000}{16.800000}\selectfont \(\displaystyle {0.0}\)}%
\end{pgfscope}%
\begin{pgfscope}%
\pgfsetbuttcap%
\pgfsetroundjoin%
\definecolor{currentfill}{rgb}{0.000000,0.000000,0.000000}%
\pgfsetfillcolor{currentfill}%
\pgfsetlinewidth{0.803000pt}%
\definecolor{currentstroke}{rgb}{0.000000,0.000000,0.000000}%
\pgfsetstrokecolor{currentstroke}%
\pgfsetdash{}{0pt}%
\pgfsys@defobject{currentmarker}{\pgfqpoint{0.000000in}{-0.048611in}}{\pgfqpoint{0.000000in}{0.000000in}}{%
\pgfpathmoveto{\pgfqpoint{0.000000in}{0.000000in}}%
\pgfpathlineto{\pgfqpoint{0.000000in}{-0.048611in}}%
\pgfusepath{stroke,fill}%
}%
\begin{pgfscope}%
\pgfsys@transformshift{2.092117in}{0.712467in}%
\pgfsys@useobject{currentmarker}{}%
\end{pgfscope}%
\end{pgfscope}%
\begin{pgfscope}%
\definecolor{textcolor}{rgb}{0.000000,0.000000,0.000000}%
\pgfsetstrokecolor{textcolor}%
\pgfsetfillcolor{textcolor}%
\pgftext[x=2.092117in,y=0.615245in,,top]{\color{textcolor}\fontsize{14.000000}{16.800000}\selectfont \(\displaystyle {0.1}\)}%
\end{pgfscope}%
\begin{pgfscope}%
\pgfsetbuttcap%
\pgfsetroundjoin%
\definecolor{currentfill}{rgb}{0.000000,0.000000,0.000000}%
\pgfsetfillcolor{currentfill}%
\pgfsetlinewidth{0.803000pt}%
\definecolor{currentstroke}{rgb}{0.000000,0.000000,0.000000}%
\pgfsetstrokecolor{currentstroke}%
\pgfsetdash{}{0pt}%
\pgfsys@defobject{currentmarker}{\pgfqpoint{0.000000in}{-0.048611in}}{\pgfqpoint{0.000000in}{0.000000in}}{%
\pgfpathmoveto{\pgfqpoint{0.000000in}{0.000000in}}%
\pgfpathlineto{\pgfqpoint{0.000000in}{-0.048611in}}%
\pgfusepath{stroke,fill}%
}%
\begin{pgfscope}%
\pgfsys@transformshift{3.353067in}{0.712467in}%
\pgfsys@useobject{currentmarker}{}%
\end{pgfscope}%
\end{pgfscope}%
\begin{pgfscope}%
\definecolor{textcolor}{rgb}{0.000000,0.000000,0.000000}%
\pgfsetstrokecolor{textcolor}%
\pgfsetfillcolor{textcolor}%
\pgftext[x=3.353067in,y=0.615245in,,top]{\color{textcolor}\fontsize{14.000000}{16.800000}\selectfont \(\displaystyle {0.2}\)}%
\end{pgfscope}%
\begin{pgfscope}%
\definecolor{textcolor}{rgb}{0.000000,0.000000,0.000000}%
\pgfsetstrokecolor{textcolor}%
\pgfsetfillcolor{textcolor}%
\pgftext[x=2.029070in,y=0.288178in,,top]{\color{textcolor}\fontsize{14.000000}{16.800000}\selectfont dropout prob}%
\end{pgfscope}%
\begin{pgfscope}%
\pgfsetbuttcap%
\pgfsetroundjoin%
\definecolor{currentfill}{rgb}{0.000000,0.000000,0.000000}%
\pgfsetfillcolor{currentfill}%
\pgfsetlinewidth{0.803000pt}%
\definecolor{currentstroke}{rgb}{0.000000,0.000000,0.000000}%
\pgfsetstrokecolor{currentstroke}%
\pgfsetdash{}{0pt}%
\pgfsys@defobject{currentmarker}{\pgfqpoint{-0.048611in}{0.000000in}}{\pgfqpoint{-0.000000in}{0.000000in}}{%
\pgfpathmoveto{\pgfqpoint{-0.000000in}{0.000000in}}%
\pgfpathlineto{\pgfqpoint{-0.048611in}{0.000000in}}%
\pgfusepath{stroke,fill}%
}%
\begin{pgfscope}%
\pgfsys@transformshift{0.705073in}{0.856363in}%
\pgfsys@useobject{currentmarker}{}%
\end{pgfscope}%
\end{pgfscope}%
\begin{pgfscope}%
\definecolor{textcolor}{rgb}{0.000000,0.000000,0.000000}%
\pgfsetstrokecolor{textcolor}%
\pgfsetfillcolor{textcolor}%
\pgftext[x=0.357623in, y=0.782497in, left, base]{\color{textcolor}\fontsize{14.000000}{16.800000}\selectfont \(\displaystyle {0.2}\)}%
\end{pgfscope}%
\begin{pgfscope}%
\pgfsetbuttcap%
\pgfsetroundjoin%
\definecolor{currentfill}{rgb}{0.000000,0.000000,0.000000}%
\pgfsetfillcolor{currentfill}%
\pgfsetlinewidth{0.803000pt}%
\definecolor{currentstroke}{rgb}{0.000000,0.000000,0.000000}%
\pgfsetstrokecolor{currentstroke}%
\pgfsetdash{}{0pt}%
\pgfsys@defobject{currentmarker}{\pgfqpoint{-0.048611in}{0.000000in}}{\pgfqpoint{-0.000000in}{0.000000in}}{%
\pgfpathmoveto{\pgfqpoint{-0.000000in}{0.000000in}}%
\pgfpathlineto{\pgfqpoint{-0.048611in}{0.000000in}}%
\pgfusepath{stroke,fill}%
}%
\begin{pgfscope}%
\pgfsys@transformshift{0.705073in}{2.295323in}%
\pgfsys@useobject{currentmarker}{}%
\end{pgfscope}%
\end{pgfscope}%
\begin{pgfscope}%
\definecolor{textcolor}{rgb}{0.000000,0.000000,0.000000}%
\pgfsetstrokecolor{textcolor}%
\pgfsetfillcolor{textcolor}%
\pgftext[x=0.357623in, y=2.221457in, left, base]{\color{textcolor}\fontsize{14.000000}{16.800000}\selectfont \(\displaystyle {0.5}\)}%
\end{pgfscope}%
\begin{pgfscope}%
\pgfsetbuttcap%
\pgfsetroundjoin%
\definecolor{currentfill}{rgb}{0.000000,0.000000,0.000000}%
\pgfsetfillcolor{currentfill}%
\pgfsetlinewidth{0.803000pt}%
\definecolor{currentstroke}{rgb}{0.000000,0.000000,0.000000}%
\pgfsetstrokecolor{currentstroke}%
\pgfsetdash{}{0pt}%
\pgfsys@defobject{currentmarker}{\pgfqpoint{-0.048611in}{0.000000in}}{\pgfqpoint{-0.000000in}{0.000000in}}{%
\pgfpathmoveto{\pgfqpoint{-0.000000in}{0.000000in}}%
\pgfpathlineto{\pgfqpoint{-0.048611in}{0.000000in}}%
\pgfusepath{stroke,fill}%
}%
\begin{pgfscope}%
\pgfsys@transformshift{0.705073in}{3.734283in}%
\pgfsys@useobject{currentmarker}{}%
\end{pgfscope}%
\end{pgfscope}%
\begin{pgfscope}%
\definecolor{textcolor}{rgb}{0.000000,0.000000,0.000000}%
\pgfsetstrokecolor{textcolor}%
\pgfsetfillcolor{textcolor}%
\pgftext[x=0.357623in, y=3.660417in, left, base]{\color{textcolor}\fontsize{14.000000}{16.800000}\selectfont \(\displaystyle {0.8}\)}%
\end{pgfscope}%
\begin{pgfscope}%
\definecolor{textcolor}{rgb}{0.000000,0.000000,0.000000}%
\pgfsetstrokecolor{textcolor}%
\pgfsetfillcolor{textcolor}%
\pgftext[x=0.288178in,y=2.295323in,,bottom,rotate=90.000000]{\color{textcolor}\fontsize{14.000000}{16.800000}\selectfont success rate}%
\end{pgfscope}%
\begin{pgfscope}%
\pgfpathrectangle{\pgfqpoint{0.705073in}{0.712467in}}{\pgfqpoint{2.647994in}{3.165712in}}%
\pgfusepath{clip}%
\pgfsetrectcap%
\pgfsetroundjoin%
\pgfsetlinewidth{1.505625pt}%
\definecolor{currentstroke}{rgb}{0.121569,0.466667,0.705882}%
\pgfsetstrokecolor{currentstroke}%
\pgfsetdash{}{0pt}%
\pgfpathmoveto{\pgfqpoint{0.831168in}{3.630358in}}%
\pgfpathlineto{\pgfqpoint{2.092117in}{2.503173in}}%
\pgfpathlineto{\pgfqpoint{3.353067in}{1.064213in}}%
\pgfusepath{stroke}%
\end{pgfscope}%
\begin{pgfscope}%
\pgfpathrectangle{\pgfqpoint{0.705073in}{0.712467in}}{\pgfqpoint{2.647994in}{3.165712in}}%
\pgfusepath{clip}%
\pgfsetrectcap%
\pgfsetroundjoin%
\pgfsetlinewidth{1.505625pt}%
\definecolor{currentstroke}{rgb}{1.000000,0.498039,0.054902}%
\pgfsetstrokecolor{currentstroke}%
\pgfsetdash{}{0pt}%
\pgfpathmoveto{\pgfqpoint{0.831168in}{3.510445in}}%
\pgfpathlineto{\pgfqpoint{2.092117in}{2.511167in}}%
\pgfpathlineto{\pgfqpoint{3.353067in}{2.399248in}}%
\pgfusepath{stroke}%
\end{pgfscope}%
\begin{pgfscope}%
\pgfpathrectangle{\pgfqpoint{0.705073in}{0.712467in}}{\pgfqpoint{2.647994in}{3.165712in}}%
\pgfusepath{clip}%
\pgfsetrectcap%
\pgfsetroundjoin%
\pgfsetlinewidth{1.505625pt}%
\definecolor{currentstroke}{rgb}{0.172549,0.627451,0.172549}%
\pgfsetstrokecolor{currentstroke}%
\pgfsetdash{}{0pt}%
\pgfpathmoveto{\pgfqpoint{0.831168in}{3.342566in}}%
\pgfpathlineto{\pgfqpoint{2.092117in}{2.942855in}}%
\pgfpathlineto{\pgfqpoint{3.353067in}{2.663057in}}%
\pgfusepath{stroke}%
\end{pgfscope}%
\begin{pgfscope}%
\pgfsetrectcap%
\pgfsetmiterjoin%
\pgfsetlinewidth{0.803000pt}%
\definecolor{currentstroke}{rgb}{0.000000,0.000000,0.000000}%
\pgfsetstrokecolor{currentstroke}%
\pgfsetdash{}{0pt}%
\pgfpathmoveto{\pgfqpoint{0.705073in}{0.856363in}}%
\pgfpathlineto{\pgfqpoint{0.705073in}{3.734283in}}%
\pgfusepath{stroke}%
\end{pgfscope}%
\begin{pgfscope}%
\pgfsetrectcap%
\pgfsetmiterjoin%
\pgfsetlinewidth{0.803000pt}%
\definecolor{currentstroke}{rgb}{0.000000,0.000000,0.000000}%
\pgfsetstrokecolor{currentstroke}%
\pgfsetdash{}{0pt}%
\pgfpathmoveto{\pgfqpoint{0.831168in}{0.712467in}}%
\pgfpathlineto{\pgfqpoint{3.353067in}{0.712467in}}%
\pgfusepath{stroke}%
\end{pgfscope}%
\begin{pgfscope}%
\definecolor{textcolor}{rgb}{0.000000,0.000000,0.000000}%
\pgfsetstrokecolor{textcolor}%
\pgfsetfillcolor{textcolor}%
\pgftext[x=2.029070in,y=3.961512in,,base]{\color{textcolor}\fontsize{16.800000}{20.160000}\selectfont Phone}%
\end{pgfscope}%
\begin{pgfscope}%
\pgfsetbuttcap%
\pgfsetmiterjoin%
\pgfsetlinewidth{0.000000pt}%
\definecolor{currentstroke}{rgb}{0.000000,0.000000,0.000000}%
\pgfsetstrokecolor{currentstroke}%
\pgfsetstrokeopacity{0.000000}%
\pgfsetdash{}{0pt}%
\pgfpathmoveto{\pgfqpoint{3.813845in}{0.712467in}}%
\pgfpathlineto{\pgfqpoint{6.461839in}{0.712467in}}%
\pgfpathlineto{\pgfqpoint{6.461839in}{3.878179in}}%
\pgfpathlineto{\pgfqpoint{3.813845in}{3.878179in}}%
\pgfpathlineto{\pgfqpoint{3.813845in}{0.712467in}}%
\pgfpathclose%
\pgfusepath{}%
\end{pgfscope}%
\begin{pgfscope}%
\pgfpathrectangle{\pgfqpoint{3.813845in}{0.712467in}}{\pgfqpoint{2.647994in}{3.165712in}}%
\pgfusepath{clip}%
\pgfsetbuttcap%
\pgfsetroundjoin%
\definecolor{currentfill}{rgb}{0.121569,0.466667,0.705882}%
\pgfsetfillcolor{currentfill}%
\pgfsetfillopacity{0.250000}%
\pgfsetlinewidth{0.000000pt}%
\definecolor{currentstroke}{rgb}{0.000000,0.000000,0.000000}%
\pgfsetstrokecolor{currentstroke}%
\pgfsetdash{}{0pt}%
\pgfpathmoveto{\pgfqpoint{3.939940in}{3.479537in}}%
\pgfpathlineto{\pgfqpoint{3.939940in}{3.253560in}}%
\pgfpathlineto{\pgfqpoint{5.200889in}{2.441108in}}%
\pgfpathlineto{\pgfqpoint{6.461839in}{0.202525in}}%
\pgfpathlineto{\pgfqpoint{6.461839in}{1.398282in}}%
\pgfpathlineto{\pgfqpoint{6.461839in}{1.398282in}}%
\pgfpathlineto{\pgfqpoint{5.200889in}{2.821053in}}%
\pgfpathlineto{\pgfqpoint{3.939940in}{3.479537in}}%
\pgfpathlineto{\pgfqpoint{3.939940in}{3.479537in}}%
\pgfpathclose%
\pgfusepath{fill}%
\end{pgfscope}%
\begin{pgfscope}%
\pgfpathrectangle{\pgfqpoint{3.813845in}{0.712467in}}{\pgfqpoint{2.647994in}{3.165712in}}%
\pgfusepath{clip}%
\pgfsetbuttcap%
\pgfsetroundjoin%
\definecolor{currentfill}{rgb}{1.000000,0.498039,0.054902}%
\pgfsetfillcolor{currentfill}%
\pgfsetfillopacity{0.250000}%
\pgfsetlinewidth{0.000000pt}%
\definecolor{currentstroke}{rgb}{0.000000,0.000000,0.000000}%
\pgfsetstrokecolor{currentstroke}%
\pgfsetdash{}{0pt}%
\pgfpathmoveto{\pgfqpoint{3.939940in}{3.252272in}}%
\pgfpathlineto{\pgfqpoint{3.939940in}{2.841287in}}%
\pgfpathlineto{\pgfqpoint{5.200889in}{2.210914in}}%
\pgfpathlineto{\pgfqpoint{6.461839in}{0.989655in}}%
\pgfpathlineto{\pgfqpoint{6.461839in}{2.241973in}}%
\pgfpathlineto{\pgfqpoint{6.461839in}{2.241973in}}%
\pgfpathlineto{\pgfqpoint{5.200889in}{2.603570in}}%
\pgfpathlineto{\pgfqpoint{3.939940in}{3.252272in}}%
\pgfpathlineto{\pgfqpoint{3.939940in}{3.252272in}}%
\pgfpathclose%
\pgfusepath{fill}%
\end{pgfscope}%
\begin{pgfscope}%
\pgfpathrectangle{\pgfqpoint{3.813845in}{0.712467in}}{\pgfqpoint{2.647994in}{3.165712in}}%
\pgfusepath{clip}%
\pgfsetbuttcap%
\pgfsetroundjoin%
\definecolor{currentfill}{rgb}{0.172549,0.627451,0.172549}%
\pgfsetfillcolor{currentfill}%
\pgfsetfillopacity{0.250000}%
\pgfsetlinewidth{0.000000pt}%
\definecolor{currentstroke}{rgb}{0.000000,0.000000,0.000000}%
\pgfsetstrokecolor{currentstroke}%
\pgfsetdash{}{0pt}%
\pgfpathmoveto{\pgfqpoint{3.939940in}{2.722558in}}%
\pgfpathlineto{\pgfqpoint{3.939940in}{2.155880in}}%
\pgfpathlineto{\pgfqpoint{5.200889in}{1.799835in}}%
\pgfpathlineto{\pgfqpoint{6.461839in}{1.149007in}}%
\pgfpathlineto{\pgfqpoint{6.461839in}{1.858783in}}%
\pgfpathlineto{\pgfqpoint{6.461839in}{1.858783in}}%
\pgfpathlineto{\pgfqpoint{5.200889in}{2.375112in}}%
\pgfpathlineto{\pgfqpoint{3.939940in}{2.722558in}}%
\pgfpathlineto{\pgfqpoint{3.939940in}{2.722558in}}%
\pgfpathclose%
\pgfusepath{fill}%
\end{pgfscope}%
\begin{pgfscope}%
\pgfsetbuttcap%
\pgfsetroundjoin%
\definecolor{currentfill}{rgb}{0.000000,0.000000,0.000000}%
\pgfsetfillcolor{currentfill}%
\pgfsetlinewidth{0.803000pt}%
\definecolor{currentstroke}{rgb}{0.000000,0.000000,0.000000}%
\pgfsetstrokecolor{currentstroke}%
\pgfsetdash{}{0pt}%
\pgfsys@defobject{currentmarker}{\pgfqpoint{0.000000in}{-0.048611in}}{\pgfqpoint{0.000000in}{0.000000in}}{%
\pgfpathmoveto{\pgfqpoint{0.000000in}{0.000000in}}%
\pgfpathlineto{\pgfqpoint{0.000000in}{-0.048611in}}%
\pgfusepath{stroke,fill}%
}%
\begin{pgfscope}%
\pgfsys@transformshift{3.939940in}{0.712467in}%
\pgfsys@useobject{currentmarker}{}%
\end{pgfscope}%
\end{pgfscope}%
\begin{pgfscope}%
\definecolor{textcolor}{rgb}{0.000000,0.000000,0.000000}%
\pgfsetstrokecolor{textcolor}%
\pgfsetfillcolor{textcolor}%
\pgftext[x=3.939940in,y=0.615245in,,top]{\color{textcolor}\fontsize{14.000000}{16.800000}\selectfont \(\displaystyle {0.0}\)}%
\end{pgfscope}%
\begin{pgfscope}%
\pgfsetbuttcap%
\pgfsetroundjoin%
\definecolor{currentfill}{rgb}{0.000000,0.000000,0.000000}%
\pgfsetfillcolor{currentfill}%
\pgfsetlinewidth{0.803000pt}%
\definecolor{currentstroke}{rgb}{0.000000,0.000000,0.000000}%
\pgfsetstrokecolor{currentstroke}%
\pgfsetdash{}{0pt}%
\pgfsys@defobject{currentmarker}{\pgfqpoint{0.000000in}{-0.048611in}}{\pgfqpoint{0.000000in}{0.000000in}}{%
\pgfpathmoveto{\pgfqpoint{0.000000in}{0.000000in}}%
\pgfpathlineto{\pgfqpoint{0.000000in}{-0.048611in}}%
\pgfusepath{stroke,fill}%
}%
\begin{pgfscope}%
\pgfsys@transformshift{5.200889in}{0.712467in}%
\pgfsys@useobject{currentmarker}{}%
\end{pgfscope}%
\end{pgfscope}%
\begin{pgfscope}%
\definecolor{textcolor}{rgb}{0.000000,0.000000,0.000000}%
\pgfsetstrokecolor{textcolor}%
\pgfsetfillcolor{textcolor}%
\pgftext[x=5.200889in,y=0.615245in,,top]{\color{textcolor}\fontsize{14.000000}{16.800000}\selectfont \(\displaystyle {0.1}\)}%
\end{pgfscope}%
\begin{pgfscope}%
\pgfsetbuttcap%
\pgfsetroundjoin%
\definecolor{currentfill}{rgb}{0.000000,0.000000,0.000000}%
\pgfsetfillcolor{currentfill}%
\pgfsetlinewidth{0.803000pt}%
\definecolor{currentstroke}{rgb}{0.000000,0.000000,0.000000}%
\pgfsetstrokecolor{currentstroke}%
\pgfsetdash{}{0pt}%
\pgfsys@defobject{currentmarker}{\pgfqpoint{0.000000in}{-0.048611in}}{\pgfqpoint{0.000000in}{0.000000in}}{%
\pgfpathmoveto{\pgfqpoint{0.000000in}{0.000000in}}%
\pgfpathlineto{\pgfqpoint{0.000000in}{-0.048611in}}%
\pgfusepath{stroke,fill}%
}%
\begin{pgfscope}%
\pgfsys@transformshift{6.461839in}{0.712467in}%
\pgfsys@useobject{currentmarker}{}%
\end{pgfscope}%
\end{pgfscope}%
\begin{pgfscope}%
\definecolor{textcolor}{rgb}{0.000000,0.000000,0.000000}%
\pgfsetstrokecolor{textcolor}%
\pgfsetfillcolor{textcolor}%
\pgftext[x=6.461839in,y=0.615245in,,top]{\color{textcolor}\fontsize{14.000000}{16.800000}\selectfont \(\displaystyle {0.2}\)}%
\end{pgfscope}%
\begin{pgfscope}%
\definecolor{textcolor}{rgb}{0.000000,0.000000,0.000000}%
\pgfsetstrokecolor{textcolor}%
\pgfsetfillcolor{textcolor}%
\pgftext[x=5.137842in,y=0.288178in,,top]{\color{textcolor}\fontsize{14.000000}{16.800000}\selectfont dropout prob}%
\end{pgfscope}%
\begin{pgfscope}%
\pgfsetbuttcap%
\pgfsetroundjoin%
\definecolor{currentfill}{rgb}{0.000000,0.000000,0.000000}%
\pgfsetfillcolor{currentfill}%
\pgfsetlinewidth{0.803000pt}%
\definecolor{currentstroke}{rgb}{0.000000,0.000000,0.000000}%
\pgfsetstrokecolor{currentstroke}%
\pgfsetdash{}{0pt}%
\pgfsys@defobject{currentmarker}{\pgfqpoint{-0.048611in}{0.000000in}}{\pgfqpoint{-0.000000in}{0.000000in}}{%
\pgfpathmoveto{\pgfqpoint{-0.000000in}{0.000000in}}%
\pgfpathlineto{\pgfqpoint{-0.048611in}{0.000000in}}%
\pgfusepath{stroke,fill}%
}%
\begin{pgfscope}%
\pgfsys@transformshift{3.813845in}{0.856363in}%
\pgfsys@useobject{currentmarker}{}%
\end{pgfscope}%
\end{pgfscope}%
\begin{pgfscope}%
\definecolor{textcolor}{rgb}{0.000000,0.000000,0.000000}%
\pgfsetstrokecolor{textcolor}%
\pgfsetfillcolor{textcolor}%
\pgftext[x=3.466395in, y=0.782497in, left, base]{\color{textcolor}\fontsize{14.000000}{16.800000}\selectfont \(\displaystyle {0.4}\)}%
\end{pgfscope}%
\begin{pgfscope}%
\pgfsetbuttcap%
\pgfsetroundjoin%
\definecolor{currentfill}{rgb}{0.000000,0.000000,0.000000}%
\pgfsetfillcolor{currentfill}%
\pgfsetlinewidth{0.803000pt}%
\definecolor{currentstroke}{rgb}{0.000000,0.000000,0.000000}%
\pgfsetstrokecolor{currentstroke}%
\pgfsetdash{}{0pt}%
\pgfsys@defobject{currentmarker}{\pgfqpoint{-0.048611in}{0.000000in}}{\pgfqpoint{-0.000000in}{0.000000in}}{%
\pgfpathmoveto{\pgfqpoint{-0.000000in}{0.000000in}}%
\pgfpathlineto{\pgfqpoint{-0.048611in}{0.000000in}}%
\pgfusepath{stroke,fill}%
}%
\begin{pgfscope}%
\pgfsys@transformshift{3.813845in}{2.295323in}%
\pgfsys@useobject{currentmarker}{}%
\end{pgfscope}%
\end{pgfscope}%
\begin{pgfscope}%
\definecolor{textcolor}{rgb}{0.000000,0.000000,0.000000}%
\pgfsetstrokecolor{textcolor}%
\pgfsetfillcolor{textcolor}%
\pgftext[x=3.466395in, y=2.221457in, left, base]{\color{textcolor}\fontsize{14.000000}{16.800000}\selectfont \(\displaystyle {0.7}\)}%
\end{pgfscope}%
\begin{pgfscope}%
\pgfsetbuttcap%
\pgfsetroundjoin%
\definecolor{currentfill}{rgb}{0.000000,0.000000,0.000000}%
\pgfsetfillcolor{currentfill}%
\pgfsetlinewidth{0.803000pt}%
\definecolor{currentstroke}{rgb}{0.000000,0.000000,0.000000}%
\pgfsetstrokecolor{currentstroke}%
\pgfsetdash{}{0pt}%
\pgfsys@defobject{currentmarker}{\pgfqpoint{-0.048611in}{0.000000in}}{\pgfqpoint{-0.000000in}{0.000000in}}{%
\pgfpathmoveto{\pgfqpoint{-0.000000in}{0.000000in}}%
\pgfpathlineto{\pgfqpoint{-0.048611in}{0.000000in}}%
\pgfusepath{stroke,fill}%
}%
\begin{pgfscope}%
\pgfsys@transformshift{3.813845in}{3.734283in}%
\pgfsys@useobject{currentmarker}{}%
\end{pgfscope}%
\end{pgfscope}%
\begin{pgfscope}%
\definecolor{textcolor}{rgb}{0.000000,0.000000,0.000000}%
\pgfsetstrokecolor{textcolor}%
\pgfsetfillcolor{textcolor}%
\pgftext[x=3.466395in, y=3.660417in, left, base]{\color{textcolor}\fontsize{14.000000}{16.800000}\selectfont \(\displaystyle {1.0}\)}%
\end{pgfscope}%
\begin{pgfscope}%
\pgfpathrectangle{\pgfqpoint{3.813845in}{0.712467in}}{\pgfqpoint{2.647994in}{3.165712in}}%
\pgfusepath{clip}%
\pgfsetrectcap%
\pgfsetroundjoin%
\pgfsetlinewidth{1.505625pt}%
\definecolor{currentstroke}{rgb}{0.121569,0.466667,0.705882}%
\pgfsetstrokecolor{currentstroke}%
\pgfsetdash{}{0pt}%
\pgfpathmoveto{\pgfqpoint{3.939940in}{3.366549in}}%
\pgfpathlineto{\pgfqpoint{5.200889in}{2.631080in}}%
\pgfpathlineto{\pgfqpoint{6.461839in}{0.800404in}}%
\pgfusepath{stroke}%
\end{pgfscope}%
\begin{pgfscope}%
\pgfpathrectangle{\pgfqpoint{3.813845in}{0.712467in}}{\pgfqpoint{2.647994in}{3.165712in}}%
\pgfusepath{clip}%
\pgfsetrectcap%
\pgfsetroundjoin%
\pgfsetlinewidth{1.505625pt}%
\definecolor{currentstroke}{rgb}{1.000000,0.498039,0.054902}%
\pgfsetstrokecolor{currentstroke}%
\pgfsetdash{}{0pt}%
\pgfpathmoveto{\pgfqpoint{3.939940in}{3.046780in}}%
\pgfpathlineto{\pgfqpoint{5.200889in}{2.407242in}}%
\pgfpathlineto{\pgfqpoint{6.461839in}{1.615814in}}%
\pgfusepath{stroke}%
\end{pgfscope}%
\begin{pgfscope}%
\pgfpathrectangle{\pgfqpoint{3.813845in}{0.712467in}}{\pgfqpoint{2.647994in}{3.165712in}}%
\pgfusepath{clip}%
\pgfsetrectcap%
\pgfsetroundjoin%
\pgfsetlinewidth{1.505625pt}%
\definecolor{currentstroke}{rgb}{0.172549,0.627451,0.172549}%
\pgfsetstrokecolor{currentstroke}%
\pgfsetdash{}{0pt}%
\pgfpathmoveto{\pgfqpoint{3.939940in}{2.439219in}}%
\pgfpathlineto{\pgfqpoint{5.200889in}{2.087473in}}%
\pgfpathlineto{\pgfqpoint{6.461839in}{1.503895in}}%
\pgfusepath{stroke}%
\end{pgfscope}%
\begin{pgfscope}%
\pgfsetrectcap%
\pgfsetmiterjoin%
\pgfsetlinewidth{0.803000pt}%
\definecolor{currentstroke}{rgb}{0.000000,0.000000,0.000000}%
\pgfsetstrokecolor{currentstroke}%
\pgfsetdash{}{0pt}%
\pgfpathmoveto{\pgfqpoint{3.813845in}{0.856363in}}%
\pgfpathlineto{\pgfqpoint{3.813845in}{3.734283in}}%
\pgfusepath{stroke}%
\end{pgfscope}%
\begin{pgfscope}%
\pgfsetrectcap%
\pgfsetmiterjoin%
\pgfsetlinewidth{0.803000pt}%
\definecolor{currentstroke}{rgb}{0.000000,0.000000,0.000000}%
\pgfsetstrokecolor{currentstroke}%
\pgfsetdash{}{0pt}%
\pgfpathmoveto{\pgfqpoint{3.939940in}{0.712467in}}%
\pgfpathlineto{\pgfqpoint{6.461839in}{0.712467in}}%
\pgfusepath{stroke}%
\end{pgfscope}%
\begin{pgfscope}%
\definecolor{textcolor}{rgb}{0.000000,0.000000,0.000000}%
\pgfsetstrokecolor{textcolor}%
\pgfsetfillcolor{textcolor}%
\pgftext[x=5.137842in,y=3.961512in,,base]{\color{textcolor}\fontsize{16.800000}{20.160000}\selectfont Rubbish}%
\end{pgfscope}%
\end{pgfpicture}%
\makeatother%
\endgroup%